    \documentclass{article}
    
    \usepackage{microtype}
    \usepackage{graphicx}
    \usepackage{subfigure}
    \usepackage{booktabs} 
    
    \usepackage{hyperref}

    \PassOptionsToPackage{table}{xcolor}
    \usepackage{colortbl}
    \usepackage[accepted]{icml2025}
    
    \usepackage{amsmath}
    \usepackage{amssymb}
    \usepackage{mathtools}
    \usepackage{amsthm}
    
    \usepackage[capitalize,noabbrev]{cleveref}
    
    \usepackage{multirow}
    \usepackage{adjustbox}
    \usepackage{listings}
    \usepackage{tcolorbox}
    \usepackage{tabularx}
    \usepackage{enumitem}

    \definecolor{background_gray}{HTML}{FAFAFA}
    \definecolor{frame_gray}{HTML}{424242}
    
    \definecolor{background_blue}{HTML}{E3F2FD}
    \definecolor{frame_blue}{HTML}{1565C0}
    
    \definecolor{background_green}{HTML}{E8F5E9}
    \definecolor{frame_green}{HTML}{2E7D32}
    
    \definecolor{background_red}{HTML}{FFEBEE}
    \definecolor{frame_red}{HTML}{C62828}
    
    \definecolor{highlight_gray}{HTML}{E0E0E0}
    \definecolor{highlight_yellow}{HTML}{FFF59D}
    
    \newcommand{\trip}[3]{%
      \begin{tabular}{@{}r@{/}r@{/}r@{}}%
        #1 & #2 & #3%
      \end{tabular}%
    }
    
    \tcbset{
      boxrule=1.5pt,
      after=\vspace{0.75cm}
    }
    
    \newcounter{snippetcounter}
    \newtcolorbox[auto counter]{snippetbox}[3][]{%
        title=\textsc{Snippet~\thesnippetcounter: #2},
        before upper={\stepcounter{snippetcounter}},  
        label={#3},  
        #1  
    }

    \theoremstyle{plain}

    \theoremstyle{definition}

    \theoremstyle{remark}

    \usepackage[textsize=tiny]{todonotes}

\begin{document}
    
    \twocolumn[
    \icmltitle{Universal Retrieval for Multimodal Trajectory Modeling}
    
    
    
    \vspace{-0.1cm}
\centerline{
\begin{tabular}{c}
\textbf{Xuan Zhang}$^{1}$ \quad
\textbf{Ziyan Jiang}$^{2}$ \quad
\textbf{Rui Meng}$^{3}$ \quad
\textbf{Yifei Leng}$^{1}$ \quad
\textbf{Zhenbang Xiao}$^{1}$ \\
\textbf{Zora Zhiruo Wang}$^{4}$ \quad
\textbf{Yanyi Shang}$^{1}$ \quad
\textbf{Dehan Kong}$^{1}$
\end{tabular}
}

\vspace{0.15cm}
\centerline{
\begin{tabular}{c}
$^{1}$iMean AI \\
$^{2}$University of California, Santa Barbara \quad
$^{3}$Google Cloud AI Research \quad
$^{4}$Carnegie Mellon University
\end{tabular}
}

    
    
    \icmlcorrespondingauthor{Dehan Kong}{dehan@imean.ai}

    
    \vskip 0.3in
    ]
    
    
    
    
    \printAffiliationsAndNotice
    
    \begin{abstract}
    Trajectory data, capturing human actions and environmental states across various modalities, holds significant potential for enhancing AI agent capabilities, particularly in GUI environments. However, how to model the representation of trajectory-level data presents a significant challenge that has not been systematically addressed amid explosive trajectory data growth. In this work, we introduce Multimodal Trajectory Retrieval, bridging the gap between universal retrieval and agent-centric trajectory modeling. We construct the Unified Agent Trajectory Dataset (UATD) from annotated demonstrations and states across diverse real-world scenarios. Based on this, we present GAE-Bench, a benchmark containing a large number of trajectory-based retrieval pairs. In addition, we propose GAE-Retriever, a multimodal retrieval framework that adopts vision-language models and incorporates optimized contrastive learning through a token selection and the GradCache mechanism. Comprehensive evaluations across multiple datasets show that GAE-Retriever consistently outperforms strong baselines in retrieval recall, highlighting its effectiveness in advancing multimodal trajectory retrieval.
    \end{abstract}
    
    \section{Introduction}
    \label{introduction}
    Human experience, meticulously recorded across diverse media like text, images, videos, and structured computational processes, forms a rich repository of knowledge known as \textit{trajectories}. These trajectories encapsulate not only the actions performed but also the environmental states in which they occurred. The vast amount of trajectory data already available in human-generated content, such as instructional videos \cite{coin} and illustrated guides \cite{show-me-more-details}, is continuously expanding through the efforts of AI agent researchers and the deployment of agent products. This wealth of experiential data offers significant value, not only for human reuse and learning but also for enhancing the intelligence in fields such as embodied intelligence \cite{embodied-mllm-retriever} and computer-use agents \cite{expel}. Given the ever-increasing volume of trajectory data, a critical question arises: \textit{How can we effectively model these trajectories to boost more advanced intelligence?}

    Prior methodologies have explored the representation of states and actions to facilitate the retrieval of optimal actions, diverging from the generation-based approach \cite{webshop}. Trajectory-level experience data have shown value within in-context reasoning paradigms \cite{regent,awm,learnact} and reinforcement training \cite{retrieval-rl,large-scale-rl-retrieval,think-before-you-act}, highlighting the rich semantic information embedded within task instructions, states, and trajectories in a latent space that can significantly aid agent inference and learning. However, trajectory-based data inherently involve substantial token consumption, making the efficient retrieval of the most pertinent trajectory data a critical challenge during the trajectory data explosion. Furthermore, state representation modeling holds immense potential for tasks such as world modeling \cite{webdreamer} and search algorithms \cite{agent-q,tree-search-agents}, presenting a viable alternative to the conventional strategies of prediction and actual interaction with environments. Despite the promise of trajectory and state representation modeling, existing studies lack a systematic evaluation of model capacities and a comprehensive understanding of how to scale these capacities with increasing data and task complexity. This work chooses GUI-based environments as an initial ground for exploration, driven by the practical value of web automation across diverse applications and the abundance of existing data resources, presented by pioneering works in this field such as Mind2Web \cite{mind2web} and WebArena \cite{webarena}.

    \begin{figure*}[ht]
        \centering
        \setlength{\abovecaptionskip}{0pt}   
        \setlength{\belowcaptionskip}{0pt}
        \centering 
        \includegraphics[width=\textwidth]{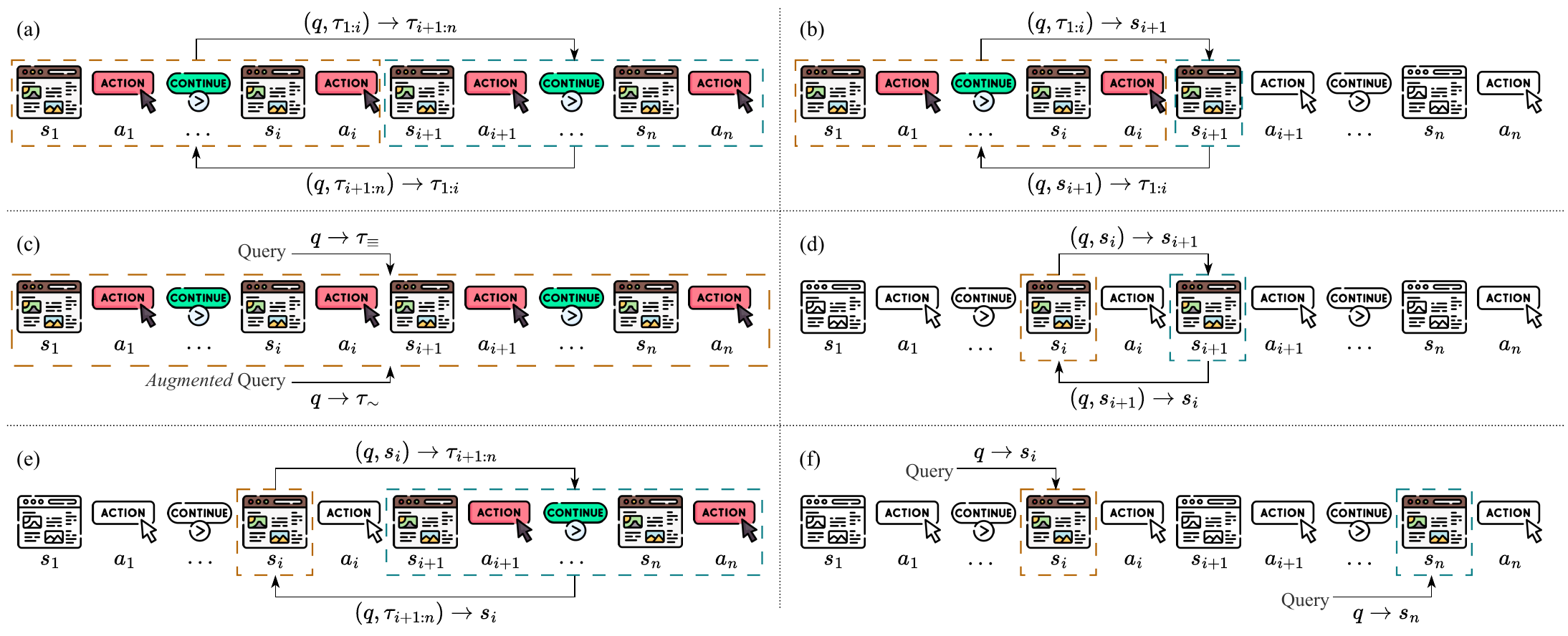}
        \vspace{-0.45cm}
        \caption{Illustration of positive pair extraction from UATD. Subfigures (a), (b), (d), and (e) depict temporal retrieval; (c) and (f) show semantic retrieval. We use $\tau_{i:j}$ to denote a subsequence of trajectory $\tau$  from state $s_i$ to action $a_j$, i.e., $\tau_{i:j} = (s_i, a_i, \ldots, s_j, a_j)$. $q$ serves as the retrieval query, composed of a task-specific instruction (shown in Table~\ref{tab:bench_overview}) along with a corresponding trajectory description. }
        \label{fig:dataset}
    \end{figure*}
    
\begin{table*}[ht]
  \centering
  \tabcolsep=5pt
  \vspace{-0.2cm}
  \setlength{\abovecaptionskip}{0pt}
  \setlength{\belowcaptionskip}{0pt}
  \caption{The overview of the GUI Agent Embedding Benchmark (GAE-Bench) series. This table summarizes the six fundamental retrieval tasks and their twelve corresponding subtasks, developed from the predefined extraction schemes in Figure~\ref{fig:dataset}. For every subtask, we present a representative instruction (see Appendix~\ref{app:prompt_list} for more instruction templates). The table also reports the number of subtask examples in GAE-Bench and the split statistics in GAE-Bench-lite: training, in-domain (IND), out-of-domain (OOD), and total instances.}
  \resizebox{\textwidth}{!}{%
    \begin{tabular}{@{}lll r r r r r@{}}
      \toprule
      \multirow{2}{*}{\textbf{Task}}
        & \multirow{2}{*}{\textbf{Subtask}}
        & \multirow{2}{*}{\textbf{Instruction}}
        & \multirow{2}{*}{\textbf{GAE-Bench}}
        & \multicolumn{4}{c}{\textbf{GAE-Bench-lite}} \\
      \cmidrule(lr){5-8}
      & & (shown 1 out of 10)
        & 
        & \textbf{Train} & \textbf{IND} & \textbf{OOD} & \textbf{Total} \\
      \midrule

      \multirow{2}{*}{1.\ $(q,\tau)\to\tau'$}
        & $(q,\tau_{1:i})\to\tau_{i+1:n}$
          & Apply the request to the previous web navigation steps to derive the next trajectory.
          & 75,046  & 40,748  &  2,523  &  2,665  & 45,936   \\
        & $(q,\tau_{i+1:n})\to\tau_{1:i}$
          & Find the previous web browsing trajectory based on the user input and the current trajectory.
          & 75,046  & 40,748  &    723  &  2,665  & 44,136   \\
      \midrule

      \multirow{2}{*}{2.\ $(q,\tau)\to s$}
        & $(q,\tau_{1:i})\to s_{i+1}$       
          & Locate the following state from the given former web navigation trajectory and the task input.
          & 75,046  & 51,661  &  3,197  &  2,665  & 57,523   \\
        & $(q,\tau_{i+1:n})\to s_i$         
          & Locate the former observation using the instruction and the upcoming web interaction trajectory.
          & 75,046  & 51,661  &  3,197  &  2,665  & 57,523   \\
      \midrule

      \multirow{2}{*}{3.\ $q\to\tau$}
        & $q\to\tau_{\equiv}$               
          & Identify the unique navigation trajectory for web agents according to the provided instruction.
          &  7,747  &  3,717  &    424  &    470  &  4,611   \\
        & $q\to\tau_{\sim}$                 
          & Retrieve the analogous interaction history for GUI agents based on the provided command.
          & 38,735  & 20,081  &    624  &  2,350  & 23,055   \\
      \midrule

      \multirow{2}{*}{4.\ $(q,s)\to s'$}
        & $(q,s_i)\to s_{i+1}$               
          & Using the provided instruction and the former state, what is the next GUI navigation state?
          & 75,046  & 68,849  &  3,532  &  2,665  & 75,046   \\
        & $(q,s_{i+1})\to s_i$               
          & Taking the description and the current state into account, search the previous web agent state.
          & 75,046  & 68,849  &  3,532  &  2,665  & 75,046   \\
      \midrule

      \multirow{2}{*}{5.\ $(q,s)\to\tau$}
        & $(q,s_i)\to\tau_{i+1:n}$           
          & Locate the next GUI navigation trajectory by applying the instruction to the previous state.
          & 75,046  & 54,124  &    734  &  2,665  & 57,523   \\
        & $(q,s_{i+1})\to\tau_{1:i}$         
          & Identify the web navigation trajectory preceding the current state according to the task.
          & 75,046  & 54,124  &    734  &  2,665  & 57,523   \\
      \midrule

      \multirow{2}{*}{6.\ $q\to s$}
        & $q\to s_i$                        
          & From the description, identify the specific navigation observation for GUI navigation.
          & 60,031  & 53,892  &  3,610  &  2,529  & 60,031   \\
        & $q\to s_n$                        
          & Search the terminal observation in the web navigation for the task instruction.
          &  7,747  &  6,502  &    775  &    470  &  7,747   \\
      \bottomrule
    \end{tabular}%
  }
  \vspace{-0.2cm}
  \label{tab:bench_overview}
\end{table*}
    
    In this paper, we propose a multimodal trajectory retrieval framework as a cornerstone for future research, consisting of three main contributions:

     1. To unify heterogeneous trajectories across web, mobile, desktop, and embodied environments, we construct the \textbf{Unified Agent Trajectory Dataset} (UATD), built from five open-source GUI agent benchmarks. UATD includes 7,747 human-annotated demonstrations and 82,793 states, covering diverse real-world use cases such as shopping, travel, business, and social media. Each trajectory features high-definition visual observations, a standardized action format, and natural language state descriptions to support customization and the development of downstream tasks.

     2. We are the first to propose the task of \textbf{Multimodal Trajectory Retrieval}, bridging the gap between universal retrieval and trajectory modeling. This task captures both temporal and semantic correlations within and across trajectories, targeting fine-grained intra-trajectory components as well as coarse-grained inter-trajectory semantic relations. To formalize the task, we design 12 extraction patterns that derive six types of retrieval samples from a single trajectory, where the query and target can be a state, a trajectory, or a subtrajectory. These include (1) text-to-state, (2) text-to-trajectory, (3) state-to-state, (4) state-to-trajectory, (5) trajectory-to-state, and (6) trajectory-to-trajectory retrieval.

    Based on this formulation, we annotate the \textbf{GUI Agent Embedding Benchmark} (GAE-Bench) by converting the Unified Agent Trajectory Dataset using the predefined extraction patterns, yielding a total of 714,628 positive retrieval pairs. To accommodate the context length limitations of current multimodal language models and enhance applicable use in downstream applications, we also release GAE-Bench-lite, a constrained version in which trajectory lengths are capped at 10 steps. GAE-Bench-lite contains 514,956 training samples, 21,805 in-domain samples, and 27,139 out-of-domain samples, with 574,222 candidates.
    
        
    3. We implement the \textbf{GUI Agent Embedding Retriever} (GAE-Retriever), a novel multimodal trajectory retrieval framework built on VLM2Vec \cite{vlm2vec}, which adopts vision-language models (VLMs), such as Qwen2-VL, as its backbone. In contrast to CLIP-based models~\cite{clip,blip,blip-2}, VLMs are pretrained on large-scale, instruction-following multimodal data and can process arbitrary-length combinations of visual and textual inputs, making them well-suited for modeling multimodal trajectories. To enable effective contrastive learning over multiple high-resolution trajectory screenshots and a large number of in-batch negatives under limited computing resources, GAE-Retriever incorporates a token selection mechanism and GradCache optimization.

    We conduct comprehensive evaluations of GAE-Retriever against multimodal backbone models, retrieval models, and trajectory planning models, showcasing its strong performance across all baselines on Recall@1/5/10. Compared to the best-performing baselines, GAE-Retriever achieves an average improvement of 10.22 points across five datasets, demonstrating the effectiveness of our training approach.

    \section{Related Work}
    \label{related work}
    \subsection{GUI Agents}
    The evolution of language models has introduced strong capabilities in tool use, environmental grounding, and complex reasoning for agentic tasks~\cite{language-agent-survey}, such as web browsing~\cite{webshop}, travel planning~\cite{travelplanner, ask-before-plan}, and societal simulation~\cite{generative-agents}. GUI navigation~\cite{gui-agent-survey}, originally derived from web-based tasks, has become an active area of research across domains including web~\cite{mind2web,webarena,visualwebarena}, mobile~\cite{android-in-the-wild,android-world}, and computer control~\cite{osworld}.
    
    To better reflect realistic conditions, GUI agents have evolved from reactive systems~\cite{pix2struct,ferret-ui-2}, 
    to proactive agents that take actions based on conversational understanding or situational reasoning. These two paradigms have led to distinct approaches: user interaction through dialogue~\cite{self-map}, environment-based reasoning~\cite{react}, and tree search~\cite{tree-search-agents}. However, real-time exploration can be inefficient and may sacrifice user experience. To reduce this overhead, some approaches retrieve reusable subroutines~\cite{awm, skillweaver} or demonstration histories from memory~\cite{rap, learnact}. Nonetheless, they perform similarity search using only textual features, neglecting richer multimodal signals.
    
    \subsection{Multimodal Retrieval}
    Recent advances in foundational vision-language models~\cite{llava, kosmos-2} have shifted research focus from generation to retrieval. Early approaches~\cite{mteb, dreamsim} primarily addressed single-modality retrieval tasks such as text-to-image and text-to-text retrieval. More recent efforts~\cite{uniir, e5-v, vlm2vec, magiclens, mm-embed} have expanded to composed and cross-modal retrieval settings, constructing large-scale benchmarks from vision-language datasets that encompass not only retrieval but also classification, captioning, and grounding tasks.
    
    Recent studies have targeted more practical domains, including visual documents~\cite{colpali}, videos~\cite{internvideo2}, and screenshots~\cite{unise}. VLM2Vec-V2~\cite{vlm2vec-v2.2}, for instance, integrates data from diverse modalities to learn universal discriminative embeddings. Despite substantial progress in multimodal retrieval, few works have explored integrating trajectory data into retrieval model training. Notably, contemporary findings \cite{autoguide, regent, embodied-mllm-retriever} highlight the potential of agent trajectories for downstream in-context planning tasks, underscoring a promising but underexplored direction in this research area.
    


    \section{Dataset}
    To build datasets for the multimodal trajectory retrieval task, a unified agent trajectory format is required for generating valid positive and negative samples. Current datasets, whether sourced from digital environments~\cite{webarena, amex} or embodied platforms~\cite{alfworld}, encode trajectories using heterogeneous structures, making consistent retrieval data extraction nontrivial. To address this challenge, we design a pipeline that converts these datasets into a standardized format, resulting in the \textbf{U}nified \textbf{A}gent \textbf{T}rajectory \textbf{D}ataset (UATD). Building upon UATD, we define 12 extraction schemes that convert each individual trajectory into labeled samples covering six core retrieval tasks, forming the basis of the \textbf{G}UI \textbf{A}gent \textbf{E}mbedding \textbf{Ben}chmark (GAE-Bench) series.

    \subsection{UATD} \label{subsec:uatd}

    To identify the essential element in trajectory representations, we model a typical agent interacting with an environment as a deterministic Markov Decision Process (MDP) $\mathcal{E} = ( \mathcal{S}, \mathcal{A}, \mathcal{O}, \mathcal{T})$, where $\mathcal{S}$ denotes the state space, $\mathcal{A}$ the action space, $\mathcal{O}$ the observation space, and $\mathcal{T}: \mathcal{S} \times \mathcal{A} \rightarrow \mathcal{S}$ is the environment transition function. A trajectory is thus represented as a sequence of state $s_i$ and action $a_i$ with $n$ steps, $\tau = (s_1, a_1, s_2, a_2, \ldots, s_n, a_n)$, where each transition satisfies $s_{i+1} = \mathcal{T}(s_i, a_i)$ and reward signals are omitted here for simplicity.
    
    Inspired by AGUVIS \cite{aguvis}, we take each observation $o_i \in \mathcal{O}$ to be the raw visual content of the interface (i.e., a screenshot).  This choice eliminates reliance on platform-specific textual representations and promotes broader generalization across visual contexts. For each action $a_i \in \mathcal{A}$, we stipulate three components: (1) operation: the name or type of the action; (2) target: the object or region within the environment on which the action is executed; (3) value: additional arguments required to perform the action. To address cross-platform discrepancies in action spaces that hinder scalability, we follow the action README from ShowUI~\cite{showui}, allowing each trajectory to be associated with its own customizable action definitions.
    
    To create the dataset under a uniform trajectory representation, we collect real-world human trajectories with their action definitions from five GUI sources: Mind2Web \cite{mind2web}, AutoWebGLM \cite{autowebglm}, WebArena \cite{webarena, visualwebarena}, WebLINX \cite{weblinx}, and GUIAct \cite{guicourse}. The raw data is manually cleaned by removing invalid trajectories and corrupted states. For sources without native visual observations (e.g., AutoWebGLM), we first complete the HTML using \texttt{gpt-4o-mini}, and then render the content via Playwright to produce screenshot-based observations. Additionally, textual descriptions are generated for each state, conditioned on the corresponding screenshot, to facilitate subsequent retrieval tasks. The prompts used for this annotation process and definition of action spaces are listed in Appendix~\ref{app:prompt_list}.
    
    Consequently, we obtain the Unified Agent Trajectory Dataset, with statistics shown in Table~\ref{tab:uatd}. In this dataset, all screenshots are preserved at their original resolution, and target elements are annotated using bounding boxes.

    \begin{table}[h]
  \scriptsize
  \vspace{-0.2cm}
  \centering
  \setlength{\abovecaptionskip}{0pt}  
  \setlength{\belowcaptionskip}{0pt}
  \caption{Statistics of the Unified Agent Trajectory Dataset (UATD). For each data source, we include the total number of tasks along with the average, minimum, maximum, and total number of states.
}
  \resizebox{\linewidth}{!}{%
    \begin{tabular}{@{}l r rrrr@{}}
      \toprule
      \multirow{2}{*}{\textbf{Source}}
        & \multirow{2}{*}{\textbf{Task}}
        & \multicolumn{4}{c}{\textbf{State}} \\
      \cmidrule(lr){3-6}
        & 
        & \textbf{Min} & \textbf{Max} & \textbf{Avg} & \textbf{Total} \\
      \midrule
      Mind2Web     & 1,468 & 2 & 31 &  6.65 &  9,621 \\
      AutoWebGLM   &   140 &   1  &    34  &    4.43 &    620 \\
      WebArena     &   201 &   1  &  26  & 6.07 &  1,221 \\
      WebLINX      &   485 &   1  &  68  &  14.19 &  6,882 \\
      GUIAct       & 5,453 & 1  & 85  & 11.82 & 64,449 \\
      \bottomrule
    \end{tabular}%
  }
  \vspace{-0.2cm}
  \label{tab:uatd}
\end{table}
    
    \subsection{GAE-Bench}
    After acquiring the unified trajectory dataset, we introduce twelve extraction patterns to generate positive retrieval pairs from each trajectory based on a query $q$, as depicted in Figure~\ref{fig:dataset}. These patterns encompass two categories of multimodal trajectory retrieval: \textbf{temporal retrieval}, which captures sequential relationships within a trajectory, and \textbf{semantic retrieval}, which targets underlying intent or function over different trajectories or states.
    
    Figure~\ref{fig:dataset} subfigures (a) and (d) detail scenarios to retrieve the next state $s_{i+1}$ or the remaining trajectory $\tau_{i+1:n}$ given a current state $s_i$ or a prefix trajectory $\tau_{1:i}$, and vice versa. Subfigures (b) and (e) demonstrate situations involving different granularities, such as identifying the correspondence between the partial trajectory $\tau_{1:i}$ and the upcoming state $s_{i+1}$, or between the present state $s_i$ and a remaining trajectory segment $\tau_{i+1:n}$. These positive pairs capture sequential relations within a trajectory in a question-answering format.

    
    Figure~\ref{fig:dataset} subfigures (c) and (f) expound two types of semantic retrieval tasks. In particular, subfigure (c) defines retrieval between a query and its reference-aligned trajectory (known as the gold trajectory), denoted as $q \to \tau_{\equiv}$, which is directly from the UATD ground truth. Conversely, subfigure (f) constructs retrieval pairs between a query and a semantically similar variant (known as the silver trajectory), denoted as $q \to \tau_{\sim}$. This represents a trajectory that preserves the same high-level intent and functional objective but varies in contextual details. For instance, the silver trajectory corresponding to “Buy a t-shirt for children on Amazon” could be “Order a laser printer on eBay”.
    
    Because generating silver trajectories is more challenging than augmenting queries, we design a three-step procedure to produce silver instructions for trajectories: (1) Identify named entities in the original query using named entity recognition (NER); (2) Generate alternative expressions for the identified entities while preserving their types; (3) Rewrite the original query after substituting the identified entities with the corresponding alternatives. Appendix~\ref{app:prompt_list} outlines the prompts for silver generation.
    
    To implement $q \to s_i$ and $q \to s_n$ in subfigure (f), state descriptions from UATD are utilized to formulate retrieval of a specific intermediate state $s_i$ and the final state $s_n$.

    \begin{table}[h]
\scriptsize
\centering
\vspace{-0.2cm}
\setlength{\abovecaptionskip}{0pt}
\setlength{\belowcaptionskip}{0pt}
\setlength{\tabcolsep}{3pt}
\caption{Distribution of GAE-Bench by retrieval tasks and data sources. This full benchmark includes 714,628 positive pairs.}
\label{tab:bench_distribution}
\begin{adjustbox}{max width=\linewidth}
\begin{tabular}{@{}lrrrrr@{}}
  \toprule
  \textbf{Retrieval Task}
    & Mind2Web & AutoWebGLM & WebArena & WebLINX & GUIAct \\
  \midrule
  $(q,\tau)\to\tau'$  
    &  16,306 &     960 &   2,040 &  12,794  & 117,992 \\
  $(q,\tau)\to s$  
    &  16,306 &     960 &   2,040 &  12,794  & 117,992 \\
  $q\to\tau$      
    &   8,808 &     840 &   1,206 &   2,910  &  32,718 \\
  $(q,s)\to s'$   
    &  16,306 &     960 &   2,040 &  12,794  & 117,992 \\
  $(q,s)\to\tau$  
    &  16,306 &     960 &   2,040 &  12,794  & 117,992 \\
  $q\to s$       
    &  10,943 &     760 &   1,305 &   6,337  &  48,433 \\
  \midrule
  \textbf{Total}
    &  84,975 &   5,440 &  10,671 &  60,423  & 553,119 \\
  \bottomrule
\end{tabular}
\end{adjustbox}
\vspace{-0.2cm}
\end{table}

\begin{table}[h]
\scriptsize
\centering
\vspace{-0.2cm}
\setlength{\abovecaptionskip}{5pt}
\setlength{\belowcaptionskip}{0pt}
\setlength{\tabcolsep}{3pt}
\caption{Distribution of GAE-Bench-lite by retrieval tasks and data sources. This compact variant includes 563,900 positive pairs.}
\label{tab:bench_lite_distribution}
\begin{adjustbox}{max width=\linewidth}
\begin{tabular}{@{}lrrrrr@{}}
  \toprule
  \textbf{Retrieval Task}
    & Mind2Web & AutoWebGLM & WebArena & WebLINX & GUIAct \\
  \midrule
  $(q,\tau)\to\tau'$  
    &  13,630 &     718 &   1,580 &   4,376 &  67,968 \\
  $(q,\tau)\to s$  
    &  14,912 &     822 &   1,796 &   7,746 &  89,770 \\
  $q\to\tau$      
    &   7,566 &     792 &   1,026 &   1,242 &  17,040 \\
  $(q,s)\to s'$   
    &  16,306 &     960 &   2,040 &  12,794 & 117,992 \\
  $(q,s)\to\tau$  
    &  14,912 &     822 &   1,796 &   7,746 &  89,770 \\
  $q\to s$       
    &  10,943 &     760 &   1,305 &   6,337 &  48,433 \\
  \midrule
  \textbf{Total}
    &  78,269 &   4,874 &   9,543 &  40,241 & 430,973 \\
  \bottomrule
\end{tabular}
\end{adjustbox}
\vspace{-0.2cm}
\end{table}

    \begin{table}[ht]
  \centering
  \scriptsize
  \setlength{\abovecaptionskip}{0pt}
  \setlength{\belowcaptionskip}{0pt}
  \setlength{\tabcolsep}{2pt}
  \vspace{-0.3cm}
  \caption{Overview of candidate sets in the GAE-Bench series. For each of the five data sources, we define three types of candidate sets (i.e., \textit{state}, \textit{trajectory}, and \textit{interval}), corresponding to retrieval targets of a single state $s_i$, a full trajectory $\tau$, and a trajectory subsequence $\tau_{i:j}$. Note that $\tau_{i:j}$ can range in length from a single state-action pair to the entire trajectory.}
  \begin{adjustbox}{max width=\textwidth}
  \begin{tabular}{@{}l r r r r r r@{}}
    \toprule
    & \textbf{Mind2Web} & \textbf{WebLINX} & \textbf{WebArena} & \textbf{GUIAct} & \textbf{AutoWebGLM} & \textbf{Total} \\
    \midrule

    \rowcolor{gray!40}\multicolumn{7}{c}{\bfseries GAE-Bench} \\
    \midrule
    \textbf{State}      &   9,475 &  5,852 &  1,104 &  42,980 &    620 & 60,031 \\
    \textbf{Trajectory} &   1,468 &    485 &    201 &   5,453 &    140 & 7,747 \\
    \textbf{Interval}   &  48,238 & 76,095 &  6,640 & 529,528 &  2,980 & 663,481 \\

    \midrule
    \rowcolor{gray!40}\multicolumn{7}{c}{\bfseries GAE-Bench-lite} \\
    \midrule
    \textbf{State}      &   9,475 &  5,852 &  1,104 &  42,980 &    620 & 60,031 \\
    \textbf{Trajectory} &   1,261 &    207 &    171 &   2,840 &    132 & 4,611 \\
    \textbf{Interval}   &  44,323 & 48,663 &  5,882 & 408,311 &  2,401 & 509,580 \\

    \midrule
    \rowcolor{gray!40}\multicolumn{7}{c}{\bfseries GAE-Bench-lite (mini)} \\
    \midrule
    \textbf{State}      &   2,842 &  2,127 &    588 &  10,530 &    349 & 16,436 \\
    \textbf{Trajectory} &     239 &     67 &     58 &     484 &     48 &    896 \\
    \textbf{Interval}   &  15,237 & 26,191 &  3,093 &  46,515 &  1,720 & 92,756 \\

    \bottomrule
  \end{tabular}
  \end{adjustbox}
  \label{tab:candidate_pool}
  \vspace{-0.3cm}
\end{table}

    Table~\ref{tab:bench_overview} outlines the extracted positive pairs organized into six retrieval tasks and twelve subtasks that comprise GAE-Bench. 
    To accommodate the limited context window of current multimodal models and maintain the feasibility of training, we also release GAE-Bench-lite, a constrained version of GAE-Bench in which each trajectory sequence contains fewer than 10 steps. To construct this version, we first randomly sample trajectories to form the out-of-domain evaluation subset. From the remaining positive pairs, we then partition the in-domain evaluation set and the training set, with approximately 90\% of the examples assigned to the training set. Detailed distributions of GAE-Bench and GAE-Bench-lite are displayed in Tables~\ref{tab:bench_distribution} and~\ref{tab:bench_lite_distribution}.
    
    Lastly, we present the statistics of the candidate sets in the GAE-Bench series in Table~\ref{tab:candidate_pool}, broken down into three types to align with different retrieval targets. To support efficient and applicable encoding during evaluation, we additionally introduce a mini partition of the GAE-Bench-lite candidate sets, referred to as GAE-Bench-lite (mini), which contains only candidates relevant to the evaluation subset.

    \section{Framework}
    This section presents the problem definition for the multimodal trajectory retrieval, along with our model,  GAE-Retriever, which extends VLM2Vec for trajectory learning.

    \begin{figure*}[ht]
        \centering
        \setlength{\abovecaptionskip}{0pt}   
        \setlength{\belowcaptionskip}{0pt}
        \centering 
        \includegraphics[width=\textwidth]{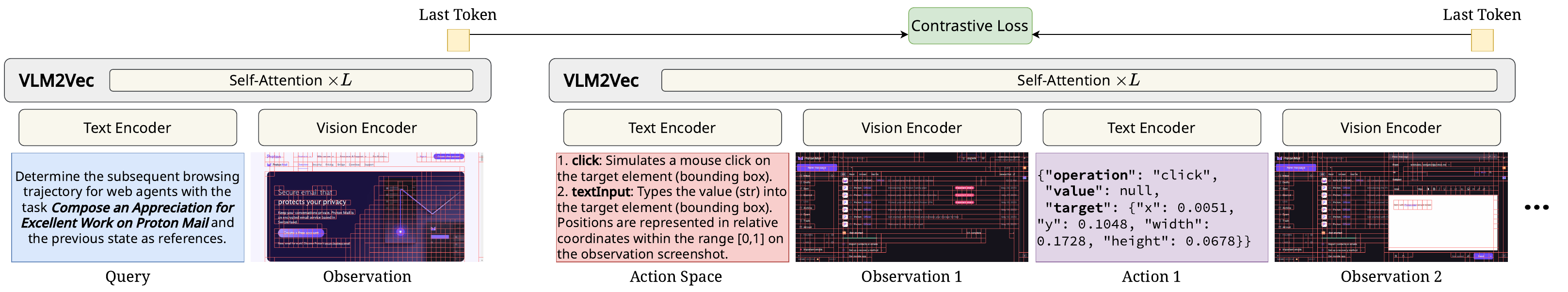}
        \vspace{-0.3cm}
        \caption{Illustration of GAE-Retriever on the subtask $(q,s_i)\to \tau_{i+1:n}$. Given a retrieval query and either a visual observation or a sequence of observation-action pairs, GAE-Retriever employs a VLM backbone to process multimodal inputs, filtering out redundant visual tokens through token selection (red blocks in observation indicate similar token clusters). Training is performed via contrastive loss between query and target sides, conditioned on task-specific instructions.}
        \label{fig:framework}
        \vspace{-0.2cm}
    \end{figure*}
    
    \subsection{Problem Definition} \label{subsec:problem_definition}
    In real-use cases, digital or embodied trajectories of users or agents may be gathered either as complete sequences $\tau$ or as partial segments, such as individual observations $s_i$ or trajectory subsequences $\tau_{i:j}$. This gives rise to diverse types of retrieval queries and candidate sets, as outlined in Table~\ref{tab:bench_overview}. The retrieval key (i.e., the input) $\mathbf{k}$ may consist of a textual query, a state, a full trajectory, or a trajectory subsequence. Similarly, the retrieval value (i.e., the target) $\mathbf{v}$ can take the form of a state, a complete sequence, or a subsequence. To handle different retrieval intentions, we formulate the structure of retrieval keys and values for model input using the following recursive grammar:
    \begin{align*}
    s_i         &\rightarrow \texttt{Observation:} [\texttt{image}_i] \\ 
    a_i         &\rightarrow [\texttt{action}_i]                  \\ 
    u_{i:j}  &\rightarrow [\texttt{action space}]\ \tau_{i:j}     \\ 
    \tau_{i:j}     &\rightarrow
      \begin{cases}
        s_i\ a_i,             & \text{if } i = j,\\
        s_i\ a_i\ \tau_{i+1:j},  & \text{if } i < j;
      \end{cases}\\
    \mathbf{k}   &\rightarrow \tilde{q}\ \mathbf{v}              \\ 
    \mathbf{v}   &\rightarrow s_i\ \mid\ u_{i:j}
    \end{align*}
    
    Here, text in teletype font denotes terminal symbols, and square brackets [$\cdot$] indicate token placeholders. A complete $n$-step trajectory is denoted by $\tau_{1:n}$, where $\tau_{i:i}$ is a state-action pair at step $i$, and $u_{i:j}$ refers to $\tau_{i:j}$ with action definitions. $\tilde{q}$ denotes $q$ combined with a task-specific instruction (see Table~\ref{tab:bench_overview}) and contextual trajectory description.
    
    Formally, we can build a multimodal trajectory retrieval model $f$ that accepts any type of key and returns any type of value as specified by $\tilde{q}$:
    \begin{align*}
        \textbf{v}^* = \operatorname*{argmax}_{\textbf{v} \in \mathcal{V}}[f(\textbf{k})^\mathsf{T} \cdot f(\mathbf{v})]
    \end{align*}
    where $\mathcal{V}$ corresponds to the heterogeneous set of candidates, $f(\cdot)$ is the retrieval function optimized via maximum dot-product similarity, and $\textbf{v}^*$ is the predicted result.
    
    \subsection{GAE-Retriever}

    We implement GAE-Retriever, a contrastive retrieval model implemented on top of VLM2Vec \cite{vlm2vec} for multimodal trajectory tasks, as portrayed in Figure~\ref{fig:framework}. The main technical challenge in modeling GAE-Retriever lies in memory constraints. Encoding these regions introduces excessive visual tokens, leading to higher inefficient computational costs and limited scalability, particularly as trajectory lengths increase. To address this, we follow prior work \cite{mixture-of-depth, showui} by constructing a UI-connected graph in RGB space to guide attention toward salient elements while skipping redundant tokens.

    In our multimodal trajectory retrieval task, obtaining hard negatives is difficult, thus making large batch sizes essential for learning high-quality embeddings through contrastive learning. However, GPU memory limitations constrain both the batch size and the number of in-batch negatives, especially since each training instance may include multiple high-resolution images, substantially increasing memory consumption. To address this, we adopt GradCache \cite{gradcache}, as in VLM2Vec, a gradient caching method that decouples backpropagation between the encoder and the contrastive loss to support larger batch training.

    To support action modeling across diverse platforms, each action in the model input is structured in JSON format with three keys: \textit{operation}, \textit{value}, and \textit{target}, accompanied by the action space definition. The \textit{target} is specified using relative bounding box coordinates. Further explanation and examples of how the input context sequence for GAE-Retriever is formulated are provided in Appendix~\ref{app:model_input_design}.

   Given a pretrained VLM, we feed $\mathbf{k}$ and $\mathbf{v}$ to obtain their embeddings, $f(\mathbf{k})$ and $f(\mathbf{v})$, by taking the final layer representation of the last token. In order to train the GAE-Retriever model, we minimize the InfoNCE loss:
    \begin{align*}
    \mathcal{L} = - \log \frac{\phi\left(f(\mathbf{k})^\mathsf{T} f(\mathbf{v}^+)\right)}{\sum_{\mathbf{v} \in \mathcal{B}} \phi\left(f(\mathbf{k})^\mathsf{T} f(\mathbf{v})\right)}
    \end{align*}
    
    where $\mathcal{B}$ is the set of in-batch candidates, $\mathbf{v}^+$ is the positive sample associated with the retrieval key $\mathbf{k}$, $t$ is the temperature parameter, and $\phi(\cdot)$ is the temperature-scaled exponential function defined as $\phi(x) = \exp\left(\frac{x}{t}\right)$.

    \section{Experiments}

    \begin{table*}[ht]
  \vspace{-0.2cm}
  \setlength{\abovecaptionskip}{5pt}   
  \setlength{\belowcaptionskip}{0pt}
  \small
  \centering
  \setlength{\tabcolsep}{1mm}
  \caption{Overall evaluation. Performance of all methods on each data source, reported in Recall@1/5/10.}
  \begin{adjustbox}{max width=\textwidth}
    \begin{tabular}{@{}l
                    rrr  rrr  rrr  rrr  rrr@{}}
      \toprule
      \multirow{2}{*}{\textbf{Method}}
        & \multicolumn{3}{c}{\textbf{Mind2Web}}
        & \multicolumn{3}{c}{\textbf{AutoWebGLM}}
        & \multicolumn{3}{c}{\textbf{WebArena}}
        & \multicolumn{3}{c}{\textbf{WebLINX}}
        & \multicolumn{3}{c}{\textbf{GUIAct}} \\
      \cmidrule(lr){2-4} \cmidrule(lr){5-7} \cmidrule(lr){8-10}
      \cmidrule(lr){11-13} \cmidrule(lr){14-16}
        & R@1 & R@5 & R@10 
        & R@1 & R@5 & R@10 
        & R@1 & R@5 & R@10 
        & R@1 & R@5 & R@10 
        & R@1 & R@5 & R@10 \\
      \midrule

      \rowcolor{gray!40}
      \multicolumn{16}{c}{\rule{0pt}{2.5ex}\bfseries Multimodal Backbone Models} \\
      \midrule
      Qwen2-VL-2B \citep{qwen2-vl} & 0.7 & 14.5 & 18.2 & 1.2 & 6.3 & 10.7 & 1.4 & 8.8 & 12.2 & 3.1 & 14.2 & 18.0 & 3.1 & 8.1 & 9.4 \\ 
      Qwen2.5-VL-3B \citep{qwen2.5-vl} & 1.0 & 7.8 & 9.7 & 0.9 & 3.8 & 6.3 & 0.7 & 6.5 & 9.9 & 3.4 & 13.0 & 15.9 & 3.0 & 7.9 & 9.5 \\
      \midrule
      \rowcolor{gray!40}
      \multicolumn{16}{c}{\rule{0pt}{2.5ex}\bfseries Multimodal Retrieval Models} \\
      \midrule
      LamRA-Ret \citep{lamra} & 1.1 & 15.1 & 19.2 & 4.9 & 15.2 & 22.8 & 2.0 & 10.3 & 14.5 & 3.4 & 16.9 & 21.5 & 4.2 & 10.4 & 12.4 \\
      ColQwen2-v1.0 \citep{colpali} & 3.2 & 22.0 & 29.9 & 3.9 & 17.7 & 26.3 & 2.9 & 13.7 & 20.0 & 4.2 & 19.6 & 25.1 & 6.2 & 15.5 & 19.2 \\
      GME-Qwen2VL-2B \citep{gme}  & 3.7 & 24.2 & 33.4 & 8.7 & 27.9 & 37.4 & 4.2 & 17.7 & 24.7 & 5.2 & 22.4 & 29.7 & 6.0 & 16.7 & 20.7 \\
      UniSE-MLLM  \citep{unise} & 0.8 & 12.6 & 16.0 & 0.3 & 4.9 & 9.5 & 1.2 & 9.0 & 12.8 & 2.9 & 14.2 & 17.6 & 3.1 & 7.9 & 9.7 \\
      VLM2Vec-Qwen2VL-2B \citep{vlm2vec} & 6.7 & 37.5 & 51.5 & 12.7 & 41.8 & 54.1 & 5.5 & 22.5 & 31.3 & 6.7 & 27.6 & 38.3 & 7.0 & 20.2 & 25.6 \\
      VLM2Vec-V2.2 \citep{vlm2vec-v2.2} & 10.2 & 44.0 & 60.1 & 15.7 & 51.2 & 67.1 & 9.1 & 29.1 & 37.8 & 10.7 & 38.4 & 50.5 & 12.2 & 33.1 & 40.6 \\
      \midrule

      \rowcolor{gray!40}
      \multicolumn{16}{c}{\rule{0pt}{2.5ex}\bfseries Multimodal Trajectory Planning Models} \\
      \midrule
      UGround-V1-2B \citep{uground}& 0.8 & 12.6 & 16.0 & 0.3 & 4.9 & 9.5 & 1.2 & 9.0 & 12.8 & 2.9 & 14.2 & 17.6 & 3.1 & 7.9 & 9.7     \\
      ShowUI-2B \citep{showui} & 1.0 & 13.3 & 17.0 & 0.8 & 6.0 & 8.2 & 1.6 & 8.5 & 11.7 & 3.3 & 13.7 & 17.3 & 3.1 & 7.9 & 9.2 \\
      UI-TARS-2B-SFT \citep{ui-tars}& 0.7 & 12.5 & 15.6 & 0.6 & 4.8 & 8.4 & 1.1 & 7.4 & 11.2 & 3.0 & 13.7 & 17.3 & 3.1 & 8.0 & 9.4 \\
      TongUI-3B \citep{tongui} & 1.3 & 9.3 & 11.4 & 0.5 & 3.9 & 7.3 & 1.5 & 7.0 & 10.5 & 3.4 & 13.8 & 17.4 & 3.0 & 8.1 & 9.7 \\
      \midrule

      \rowcolor{gray!40}
      \multicolumn{16}{c}{\rule{0pt}{2.5ex}\bfseries Multimodal Trajectory Retrieval Models} \\
      \midrule
      \textbf{GAE-Retriever (Ours)} & \textbf{15.0} & \textbf{50.7} & \textbf{67.6} & \textbf{22.1} & \textbf{63.6} & \textbf{76.3} & \textbf{10.3} & \textbf{31.7} & \textbf{44.1} & \textbf{13.7} & \textbf{41.7} & \textbf{54.1} & \textbf{25.7} & \textbf{59.2} & \textbf{67.9} \\
      \bottomrule
    \end{tabular}
  \end{adjustbox}
  \vspace{-0.3cm}
  \label{tab:main_results}
\end{table*}
    
    This section showcases our experimental setup and analyzes both aggregate and task-specific level results. Subtask-level performance details are available in Appendix~\ref{app:subtask_performance}.

    \subsection{Experimental Setups}\label{subsec:experimental_setups}

    \paragraph{Baselines}  We evaluate three groups of baseline models in this study: (1) \textbf{Multimodal Backbone Models}: We adopt widely used multimodal LLMs, including Qwen2-VL~\cite{qwen2-vl} and Qwen2.5-VL~\cite{qwen2.5-vl}, both of which support multi-frame image and video inference effectively. To ensure fair comparison and reduce GPU memory usage, we include only methods fine-tuned on Qwen2-VL-2B and Qwen2.5-VL-3B. (2) \textbf{Multimodal Retrieval Models}: This group includes text-and-image retrieval models (Lamra-Ret~\cite{lamra}, VLM2Vec-Qwen2VL-2B~\cite{vlm2vec}), visual document models (Colqwen~\cite{colpali}), screenshot retrieval models (UniSE-MLLM~\cite{unise}), and mixed-modality pretrained models (GME~\cite{gme}, VLM2Vec-v2.2~\cite{vlm2vec-v2.2}). These models are trained on diverse data types, including text, images, videos, documents, screenshots, and fused modalities, covering tasks such as grounding, classification, question answering, and retrieval to enable comprehensive evaluation. (3) \textbf{Multimodal Trajectory Planning Models}: This group includes four agent models: UGround-V1-2B~\cite{uground}, ShowUI-2B~\cite{showui}, UI-TARS-2B-SFT~\cite{ui-tars}, and TongUI-3B~\cite{tongui}, trained on web, mobile, and desktop interaction data, specifically for visual grounding and GUI navigation tasks. All baseline models are evaluated in a zero-shot setting.
    
    \paragraph{Implementation Details} We use the OpenAI closed-source model \texttt{gpt-4o-mini-2024-07-18} in our data annotation process. The retrieval model is trained with Qwen2-VL-Instruct using LoRA with a rank of 8 using 16 NVIDIA H800 GPUs. GradCache is employed during training with a sub-batch size of 1 and a total accumulated batch size of 2,048. 
    We train the model for 256 steps 
    on GAE-Bench-lite, for 1,044 GPU hours. The training configuration includes a learning rate of $5 \times 10^{-5}$, a 5\% warm-up ratio, and an interleaved batch size of 0.2.
    The maximum token length is set to 65,536 for both training and evaluation. Since the UI-graph-based token selection introduces no additional learnable parameters and preserves consistent positional relationships within the full token sequence regardless of its activation, we enable it only during training, applying the optimal mask ratio of 0.5~\cite{showui} across all transformer layers. 
    Evaluation is performed with a batch size of 6 using 8 H800 GPUs on GAE-Bench-lite with the mini candidate sets, requiring 22.5 GPU hours.

    \begin{figure*}[t]
        \centering
        \setlength{\abovecaptionskip}{0pt}   
        \setlength{\belowcaptionskip}{0pt}
        \centering 
        \includegraphics[width=\textwidth]{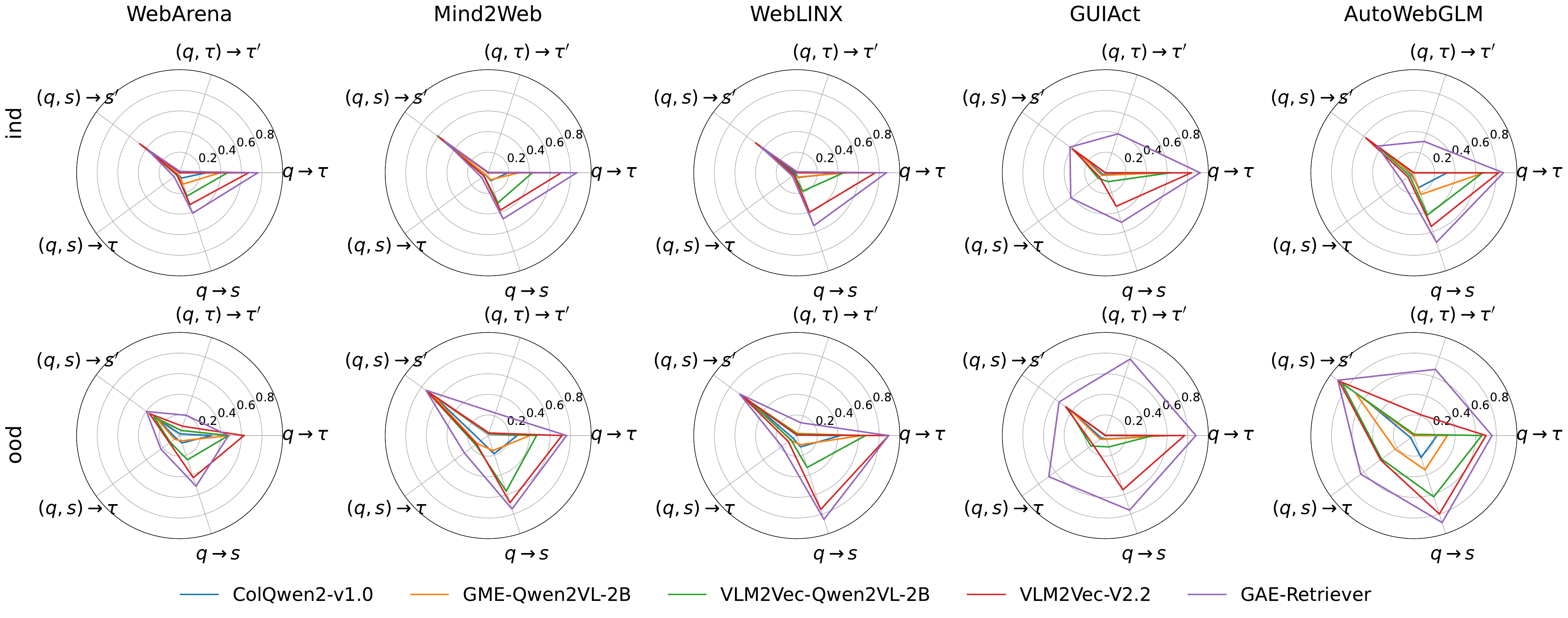}
        \vspace{-0.3cm}
        \caption{Per-Task Evaluation Results. Recall@5 is reported for five selected models across various retrieval tasks and datasets under in-domain (ind) and out-of-domain (ood) scenarios.}
        \label{fig:per-task-evaluation}
    \vspace{-0.2cm}
    \end{figure*}
    
    \subsection{Overall Evaluation}
    Table~\ref{tab:main_results} summarizes Recall@1/5/10 of all baseline models across five datasets.  Our proposed method, \textit{GAE-Retriever}, surpasses all baselines by a substantial margin on all evaluation metrics. Its consistently outstanding performance over all datasets reflects the robustness and reliability of the proposed benchmark for evaluating trajectory retrieval.
    \paragraph{Multimodal Backbone Models} The two backbone models, Qwen2-VL-2B and Qwen2.5-VL-3B, display relatively low retrieval performance compared to their understanding and reasoning skills. Their Recall@1 scores remain below 4.0 across all datasets, indicating the need for task-specific adaptation to develop effective retrieval abilities. Interestingly, Qwen2.5-VL-3B performs slightly worse than the smaller Qwen2-VL-2B on most Recall@K metrics, underscoring a substantial gap between current multimodal LLM pretraining and the demands of trajectory-based retrieval.
    
    \paragraph{Multimodal Retrieval Models} Among retrieval-focused baselines, VLM2Vec-V2.2 delivers both the highest-scoring and most stable performance across all benchmarks. It consistently surpasses its predecessor, VLM2Vec-Qwen2VL-2B, as a result of the integration of visual document and video data during training. This improvement highlights the importance of incorporating diverse multimodal data sources beyond basic image and text pairs.
    
    Models in the VLM2Vec family consistently outperform other approaches, indicating that training on comprehensive fused-modality data with interleaved batches plays a critical role in retrieval effectiveness. In contrast, LamRA-Ret, which is trained only on text and image datasets, lags behind models exposed to structured visual content like ColQwen2-v1.0 and GME-Qwen2VL-2B. While LamRA-Ret still exceeds the backbone baselines, the results confirm that text-image training is helpful, but insufficient on its own for trajectory-oriented retrieval.
    
    Between the two document-pretrained models, GME-Qwen2VL-2B achieves better results than ColQwen2-v1.0, likely due to its exposure to a broader range of visual and textual inputs beyond documents. UniSE-MLLM, finetuned on screenshots, records the weakest outcomes, even underperforming the backbone models in some cases. This reveals that screenshot-specific pretraining may lack the flexibility and generalization required for open-domain retrieval.
    
    \paragraph{Multimodal Trajectory Planning Models} Methods in this group (e.g., ShowUI-2B, TongUI-3B, UGround-V1-2B) show no significant performance improvement over the backbone models. Their modest retrieval results imply that capabilities in planning or grounding generation do not directly translate to multimodal retrieval proficiency. Additionally, the difference in model size, i.e., 2B vs. 3B, appears to have minimal impact in our experiment.
    \paragraph{Trajectory Retrieval Models} GAE-Retriever achieves the best results across all datasets and evaluation metrics. Notably, it reaches a Recall@10 of 76.3 on AutoWebGLM and 67.9 on GUIAct, with Recall@1 consistently above 10.0 across all benchmarks. Compared to the strongest baseline, VLM2Vec-V2.2, GAE-Retriever improves Recall@1 by up to 7.1 points on GUIAct and 6.4 points on AutoWebGLM. These gains validate the design of our training strategy.
    
    \subsection{Per-Task Evaluation}
    Figure~\ref{fig:per-task-evaluation} presents Recall@5 results broken down by retrieval task across multiple datasets for five representative models: ColQwen2-v1.0, GME-Qwen2VL-2B, VLM2Vec-Qwen2VL-2B, VLM2Vec-V2.2, and GAE-Retriever, under both in-domain and out-of-domain settings. As depicted, all models achieve their best results on the $q \to \tau$ and $q \to s$ tasks, indicating that semantic retrieval tasks are relatively simple. On the contrary, performance declines significantly on $(q,\tau) \to \tau'$ and $(q,s) \to \tau$, which can be attributed to the absence of similar supervision in existing benchmarks.

    GAE-Retriever attains the highest scores on nearly every task over all datasets, especially evident on GUIAct and AutoWebGLM. Notably, it performs even better in out-of-domain scenarios than in-domain ones for tasks like $q \to s$, $(q,\tau) \to \tau'$, and $(q,s) \to \tau$, showcasing the robust generalization capability of our proposed framework.
    
    Among baseline methods, VLM2Vec-V2.2 stands out as the most competitive, showing solid results across most tasks. It even surpasses GAE-Retriever on $q \to \tau$ in the out-of-domain WebArena, which can be attributed to its enhanced transfer capabilities from video pretraining. However, it consistently falls short of GAE-Retriever on more complex tasks, highlighting the necessity of retrieval models optimized for the multimodal trajectory retrieval task.
    
    VLM2Vec-Qwen2VL-2B and GME-Qwen2VL-2B achieve moderate results but remain behind the top two models. It can be observed that they outperform the document-only pretrained ColQwen2-v1.0, proving the importance of training on diverse domains. The relatively poor performance of ColQwen2-v1.0 suggests its limited adaptability beyond documents. Meanwhile, GME-Qwen2VL-2B's gap with its VLM2Vec counterpart points to deficiencies in training.
    
    Overall, the consistent advantage of GAE-Retriever reinforces the value of trajectory-based modeling and retrieval training for building versatile multimodal retrieval systems.

    \section{Conclusion}

    In this work, we introduce the task of Multimodal Trajectory Retrieval and present UATD, a unified-format dataset of real-world, GUI-based agent trajectories. Building on this, we construct two standardized benchmarks, GAE-Bench and GAE-Bench-lite, to support evaluation. To address the challenges of long multimodal sequences, we propose GAE-Retriever, a VLM2Vec-based retrieval framework enhanced with token selection and GradCache for efficient contrastive learning. Experimental results show that GAE-Retriever achieves the best performance across all five environments. Compared to the strongest retrieval baseline (VLM2Vec-V2.2), it improves Recall@1 by up to 12.9 points. 
    More importantly, as an ongoing effort, our recipe for constructing a multimodal trajectory retrieval system lays the groundwork for future investigations into retrieval-based context learning, reinforcement learning, and world modeling.

    \section*{Impact Statement}
    
    This position paper transforms open-source GUI datasets into a unified trajectory representation for multimodal retrieval. This resource may accelerate future research; however, potential concerns regarding user privacy and unintended misuse still exist when using public data. Addressing these risks requires thoughtful design and transparent practices to ensure responsible and ethical deployment.

    
    \bibliography{reference}

\begin{thebibliography}{64}
\providecommand{\natexlab}[1]{#1}
\providecommand{\url}[1]{\texttt{#1}}
\expandafter\ifx\csname urlstyle\endcsname\relax
  \providecommand{\doi}[1]{doi: #1}\else
  \providecommand{\doi}{doi: \begingroup \urlstyle{rm}\Url}\fi

\bibitem[Bai et~al.(2025)Bai, Chen, Liu, Wang, Ge, Song, Dang, Wang, Wang,
  Tang, et~al.]{qwen2.5-vl}
Bai, S., Chen, K., Liu, X., Wang, J., Ge, W., Song, S., Dang, K., Wang, P.,
  Wang, S., Tang, J., et~al.
\newblock Qwen2. 5-vl technical report.
\newblock \emph{ArXiv preprint}, abs/2502.13923, 2025.
\newblock URL \url{https://arxiv.org/abs/2502.13923}.

\bibitem[Chai et~al.(2024)Chai, Huang, Niu, Xiao, Liu, Zhang, Gao, Ren, and
  Li]{amex}
Chai, Y., Huang, S., Niu, Y., Xiao, H., Liu, L., Zhang, D., Gao, P., Ren, S.,
  and Li, H.
\newblock Amex: Android multi-annotation expo dataset for mobile gui agents.
\newblock \emph{ArXiv preprint}, abs/2407.17490, 2024.
\newblock URL \url{https://arxiv.org/abs/2407.17490}.

\bibitem[Chen et~al.(2024)Chen, Cui, Hu, Qin, Fang, Zhao, Wang, Liu, Chen, Huo,
  et~al.]{guicourse}
Chen, W., Cui, J., Hu, J., Qin, Y., Fang, J., Zhao, Y., Wang, C., Liu, J.,
  Chen, G., Huo, Y., et~al.
\newblock Guicourse: From general vision language models to versatile gui
  agents.
\newblock \emph{ArXiv preprint}, abs/2406.11317, 2024.
\newblock URL \url{https://arxiv.org/abs/2406.11317}.

\bibitem[Deng et~al.(2023)Deng, Gu, Zheng, Chen, Stevens, Wang, Sun, and
  Su]{mind2web}
Deng, X., Gu, Y., Zheng, B., Chen, S., Stevens, S., Wang, B., Sun, H., and Su,
  Y.
\newblock Mind2web: Towards a generalist agent for the web.
\newblock In \emph{Advances in Neural Information Processing Systems 36: Annual
  Conference on Neural Information Processing Systems 2023, NeurIPS 2023, New
  Orleans, LA, USA, December 10 - 16, 2023}, 2023.

\bibitem[Deng et~al.(2024)Deng, Zhang, Zhang, Yuan, Ng, and Chua]{self-map}
Deng, Y., Zhang, X., Zhang, W., Yuan, Y., Ng, S.-K., and Chua, T.-S.
\newblock On the multi-turn instruction following for conversational web
  agents.
\newblock \emph{ArXiv preprint}, abs/2402.15057, 2024.
\newblock URL \url{https://arxiv.org/abs/2402.15057}.

\bibitem[Faysse et~al.(2024)Faysse, Sibille, Wu, Omrani, Viaud, Hudelot, and
  Colombo]{colpali}
Faysse, M., Sibille, H., Wu, T., Omrani, B., Viaud, G., Hudelot, C., and
  Colombo, P.
\newblock Colpali: Efficient document retrieval with vision language models.
\newblock In \emph{The Thirteenth International Conference on Learning
  Representations}, 2024.

\bibitem[Fu et~al.(2023)Fu, Tamir, Sundaram, Chai, Zhang, Dekel, and
  Isola]{dreamsim}
Fu, S., Tamir, N., Sundaram, S., Chai, L., Zhang, R., Dekel, T., and Isola, P.
\newblock Dreamsim: Learning new dimensions of human visual similarity using
  synthetic data.
\newblock In \emph{Advances in Neural Information Processing Systems 36: Annual
  Conference on Neural Information Processing Systems 2023, NeurIPS 2023, New
  Orleans, LA, USA, December 10 - 16, 2023}, 2023.

\bibitem[Fu et~al.(2024)Fu, Kim, Kim, Sohn, Logeswaran, Bae, and
  Lee]{autoguide}
Fu, Y., Kim, D., Kim, J., Sohn, S., Logeswaran, L., Bae, K., and Lee, H.
\newblock Autoguide: Automated generation and selection of context-aware
  guidelines for large language model agents.
\newblock In \emph{Advances in Neural Information Processing Systems 38: Annual
  Conference on Neural Information Processing Systems 2024, NeurIPS 2024,
  Vancouver, BC, Canada, December 10 - 15, 2024}, 2024.

\bibitem[Gao et~al.(2021)Gao, Zhang, Han, and Callan]{gradcache}
Gao, L., Zhang, Y., Han, J., and Callan, J.
\newblock Scaling deep contrastive learning batch size under memory limited
  setup.
\newblock In \emph{Proceedings of the 6th Workshop on Representation Learning
  for NLP (RepL4NLP-2021)}, pp.\  316--321, 2021.
\newblock \doi{10.18653/v1/2021.repl4nlp-1.31}.
\newblock URL \url{https://aclanthology.org/2021.repl4nlp-1.31}.

\bibitem[Gou et~al.(2024)Gou, Wang, Zheng, Xie, Chang, Shu, Sun, and
  Su]{uground}
Gou, B., Wang, R., Zheng, B., Xie, Y., Chang, C., Shu, Y., Sun, H., and Su, Y.
\newblock Navigating the digital world as humans do: Universal visual grounding
  for gui agents.
\newblock \emph{ArXiv preprint}, abs/2410.05243, 2024.
\newblock URL \url{https://arxiv.org/abs/2410.05243}.

\bibitem[Goyal et~al.(2022)Goyal, Friesen, Banino, Weber, Ke, Badia, Guez,
  Mirza, Humphreys, Konyushkova, Valko, Osindero, Lillicrap, Heess, and
  Blundell]{retrieval-rl}
Goyal, A., Friesen, A.~L., Banino, A., Weber, T., Ke, N.~R., Badia, A.~P.,
  Guez, A., Mirza, M., Humphreys, P.~C., Konyushkova, K., Valko, M., Osindero,
  S., Lillicrap, T.~P., Heess, N., and Blundell, C.
\newblock Retrieval-augmented reinforcement learning.
\newblock In \emph{International Conference on Machine Learning, {ICML} 2022,
  17-23 July 2022, Baltimore, Maryland, {USA}}, volume 162 of \emph{Proceedings
  of Machine Learning Research}, pp.\  7740--7765, 2022.
\newblock URL \url{https://proceedings.mlr.press/v162/goyal22a.html}.

\bibitem[Gu et~al.(2024)Gu, Zhang, Ning, Zheng, Gou, Xue, Chang, Srivastava,
  Xie, Qi, et~al.]{webdreamer}
Gu, Y., Zhang, K., Ning, Y., Zheng, B., Gou, B., Xue, T., Chang, C.,
  Srivastava, S., Xie, Y., Qi, P., et~al.
\newblock Is your llm secretly a world model of the internet? model-based
  planning for web agents.
\newblock \emph{ArXiv preprint}, abs/2411.06559, 2024.
\newblock URL \url{https://arxiv.org/abs/2411.06559}.

\bibitem[Humphreys et~al.(2022)Humphreys, Guez, Tieleman, Sifre, Weber, and
  Lillicrap]{large-scale-rl-retrieval}
Humphreys, P.~C., Guez, A., Tieleman, O., Sifre, L., Weber, T., and Lillicrap,
  T.~P.
\newblock Large-scale retrieval for reinforcement learning.
\newblock In \emph{Advances in Neural Information Processing Systems 35: Annual
  Conference on Neural Information Processing Systems 2022, NeurIPS 2022, New
  Orleans, LA, USA, November 28 - December 9, 2022}, 2022.

\bibitem[Jiang et~al.(2024)Jiang, Song, Zhang, Huang, Deng, Sun, Zhang, Wang,
  and Zhuang]{e5-v}
Jiang, T., Song, M., Zhang, Z., Huang, H., Deng, W., Sun, F., Zhang, Q., Wang,
  D., and Zhuang, F.
\newblock E5-v: Universal embeddings with multimodal large language models.
\newblock \emph{ArXiv preprint}, abs/2407.12580, 2024.
\newblock URL \url{https://arxiv.org/abs/2407.12580}.

\bibitem[Jiang et~al.(2025)Jiang, Meng, Yang, Yavuz, Zhou, and Chen]{vlm2vec}
Jiang, Z., Meng, R., Yang, X., Yavuz, S., Zhou, Y., and Chen, W.
\newblock {VLM}2vec: Training vision-language models for massive multimodal
  embedding tasks.
\newblock In \emph{The Thirteenth International Conference on Learning
  Representations}, 2025.
\newblock URL \url{https://openreview.net/forum?id=TE0KOzWYAF}.

\bibitem[Kagaya et~al.(2024)Kagaya, Yuan, Lou, Karlekar, Pranata, Kinose,
  Oguri, Wick, and You]{rap}
Kagaya, T., Yuan, T.~J., Lou, Y., Karlekar, J., Pranata, S., Kinose, A., Oguri,
  K., Wick, F., and You, Y.
\newblock Rap: Retrieval-augmented planning with contextual memory for
  multimodal llm agents.
\newblock \emph{ArXiv preprint}, abs/2402.03610, 2024.
\newblock URL \url{https://arxiv.org/abs/2402.03610}.

\bibitem[Kang et~al.(2024)Kang, Laroche, Yuan, Trischler, Liu, and
  Fu]{think-before-you-act}
Kang, J., Laroche, R., Yuan, X., Trischler, A., Liu, X., and Fu, J.
\newblock Think before you act: Decision transformers with working memory.
\newblock In \emph{Forty-first International Conference on Machine Learning,
  {ICML} 2024, Vienna, Austria, July 21-27, 2024}, 2024.
\newblock URL \url{https://openreview.net/forum?id=PSQ5Z920M8}.

\bibitem[Koh et~al.(2024{\natexlab{a}})Koh, Lo, Jang, Duvvur, Lim, Huang,
  Neubig, Zhou, Salakhutdinov, and Fried]{visualwebarena}
Koh, J.~Y., Lo, R., Jang, L., Duvvur, V., Lim, M.~C., Huang, P.-Y., Neubig, G.,
  Zhou, S., Salakhutdinov, R., and Fried, D.
\newblock Visualwebarena: Evaluating multimodal agents on realistic visual web
  tasks.
\newblock \emph{ArXiv preprint}, abs/2401.13649, 2024{\natexlab{a}}.
\newblock URL \url{https://arxiv.org/abs/2401.13649}.

\bibitem[Koh et~al.(2024{\natexlab{b}})Koh, McAleer, Fried, and
  Salakhutdinov]{tree-search-agents}
Koh, J.~Y., McAleer, S., Fried, D., and Salakhutdinov, R.
\newblock Tree search for language model agents.
\newblock \emph{ArXiv preprint}, abs/2407.01476, 2024{\natexlab{b}}.
\newblock URL \url{https://arxiv.org/abs/2407.01476}.

\bibitem[Lai et~al.(2024)Lai, Liu, Iong, Yao, Chen, Shen, Yu, Zhang, Zhang,
  Dong, and Tang]{autowebglm}
Lai, H., Liu, X., Iong, I.~L., Yao, S., Chen, Y., Shen, P., Yu, H., Zhang, H.,
  Zhang, X., Dong, Y., and Tang, J.
\newblock Autowebglm: {A} large language model-based web navigating agent.
\newblock In \emph{Proceedings of the 30th {ACM} {SIGKDD} Conference on
  Knowledge Discovery and Data Mining, {KDD} 2024, Barcelona, Spain, August
  25-29, 2024}, pp.\  5295--5306, 2024.
\newblock \doi{10.1145/3637528.3671620}.
\newblock URL \url{https://doi.org/10.1145/3637528.3671620}.

\bibitem[Lee et~al.(2023)Lee, Joshi, Turc, Hu, Liu, Eisenschlos, Khandelwal,
  Shaw, Chang, and Toutanova]{pix2struct}
Lee, K., Joshi, M., Turc, I.~R., Hu, H., Liu, F., Eisenschlos, J.~M.,
  Khandelwal, U., Shaw, P., Chang, M., and Toutanova, K.
\newblock Pix2struct: Screenshot parsing as pretraining for visual language
  understanding.
\newblock In \emph{International Conference on Machine Learning, {ICML} 2023,
  23-29 July 2023, Honolulu, Hawaii, {USA}}, volume 202 of \emph{Proceedings of
  Machine Learning Research}, pp.\  18893--18912, 2023.
\newblock URL \url{https://proceedings.mlr.press/v202/lee23g.html}.

\bibitem[Li et~al.(2022)Li, Li, Xiong, and Hoi]{blip}
Li, J., Li, D., Xiong, C., and Hoi, S. C.~H.
\newblock {BLIP:} bootstrapping language-image pre-training for unified
  vision-language understanding and generation.
\newblock In \emph{International Conference on Machine Learning, {ICML} 2022,
  17-23 July 2022, Baltimore, Maryland, {USA}}, volume 162 of \emph{Proceedings
  of Machine Learning Research}, pp.\  12888--12900, 2022.
\newblock URL \url{https://proceedings.mlr.press/v162/li22n.html}.

\bibitem[Li et~al.(2023)Li, Li, Savarese, and Hoi]{blip-2}
Li, J., Li, D., Savarese, S., and Hoi, S. C.~H.
\newblock {BLIP-2:} bootstrapping language-image pre-training with frozen image
  encoders and large language models.
\newblock In \emph{International Conference on Machine Learning, {ICML} 2023,
  23-29 July 2023, Honolulu, Hawaii, {USA}}, volume 202 of \emph{Proceedings of
  Machine Learning Research}, pp.\  19730--19742, 2023.
\newblock URL \url{https://proceedings.mlr.press/v202/li23q.html}.

\bibitem[Li et~al.(2024)Li, You, Zhang, Feng, Agrawal, Li, Moorthy, Nichols,
  Yang, and Gan]{ferret-ui-2}
Li, Z., You, K., Zhang, H., Feng, D., Agrawal, H., Li, X., Moorthy, M. P.~S.,
  Nichols, J., Yang, Y., and Gan, Z.
\newblock Ferret-ui 2: Mastering universal user interface understanding across
  platforms.
\newblock \emph{ArXiv preprint}, abs/2410.18967, 2024.
\newblock URL \url{https://arxiv.org/abs/2410.18967}.

\bibitem[Lin et~al.(2024{\natexlab{a}})Lin, Li, Gao, Yang, Wu, Bai, Lei, Wang,
  and Shou]{showui}
Lin, K.~Q., Li, L., Gao, D., Yang, Z., Wu, S., Bai, Z., Lei, W., Wang, L., and
  Shou, M.~Z.
\newblock Showui: One vision-language-action model for gui visual agent.
\newblock \emph{ArXiv preprint}, abs/2411.17465, 2024{\natexlab{a}}.
\newblock URL \url{https://arxiv.org/abs/2411.17465}.

\bibitem[Lin et~al.(2024{\natexlab{b}})Lin, Lee, Shoeybi, Lin, Catanzaro, and
  Ping]{mm-embed}
Lin, S.-C., Lee, C., Shoeybi, M., Lin, J., Catanzaro, B., and Ping, W.
\newblock Mm-embed: Universal multimodal retrieval with multimodal llms.
\newblock \emph{ArXiv preprint}, abs/2411.02571, 2024{\natexlab{b}}.
\newblock URL \url{https://arxiv.org/abs/2411.02571}.

\bibitem[Liu et~al.(2025{\natexlab{a}})Liu, Zhao, Liu, Chen, Chai, Ren, Wang,
  He, and Meng]{learnact}
Liu, G., Zhao, P., Liu, L., Chen, Z., Chai, Y., Ren, S., Wang, H., He, S., and
  Meng, W.
\newblock Learnact: Few-shot mobile gui agent with a unified demonstration
  benchmark.
\newblock \emph{ArXiv preprint}, abs/2504.13805, 2025{\natexlab{a}}.
\newblock URL \url{https://arxiv.org/abs/2504.13805}.

\bibitem[Liu et~al.(2023)Liu, Li, Wu, and Lee]{llava}
Liu, H., Li, C., Wu, Q., and Lee, Y.~J.
\newblock Visual instruction tuning.
\newblock In \emph{Advances in Neural Information Processing Systems 36: Annual
  Conference on Neural Information Processing Systems 2023, NeurIPS 2023, New
  Orleans, LA, USA, December 10 - 16, 2023}, 2023.

\bibitem[Liu et~al.(2024)Liu, Chen, Cai, Jiang, Hu, Yao, Wang, and Xie]{lamra}
Liu, Y., Chen, P., Cai, J., Jiang, X., Hu, Y., Yao, J., Wang, Y., and Xie, W.
\newblock Lamra: Large multimodal model as your advanced retrieval assistant.
\newblock \emph{ArXiv preprint}, abs/2412.01720, 2024.
\newblock URL \url{https://arxiv.org/abs/2412.01720}.

\bibitem[Liu et~al.(2025{\natexlab{b}})Liu, Liang, Zhou, Liu, and Lian]{unise}
Liu, Z., Liang, Z., Zhou, J., Liu, Z., and Lian, D.
\newblock Any information is just worth one single screenshot: Unifying search
  with visualized information retrieval.
\newblock \emph{ArXiv preprint}, abs/2502.11431, 2025{\natexlab{b}}.
\newblock URL \url{https://arxiv.org/abs/2502.11431}.

\bibitem[Lu et~al.(2024)Lu, Kasner, and Reddy]{weblinx}
Lu, X.~H., Kasner, Z., and Reddy, S.
\newblock Weblinx: Real-world website navigation with multi-turn dialogue.
\newblock In \emph{Forty-first International Conference on Machine Learning,
  {ICML} 2024, Vienna, Austria, July 21-27, 2024}, 2024.
\newblock URL \url{https://openreview.net/forum?id=mUSPhG4uDW}.

\bibitem[Meng et~al.(2025)Meng, Jiang, Liu, Su, Yang, Fu, Qin, Chen, Xu, Xiong,
  Zhou, Chen, and Yavuz]{vlm2vec-v2.2}
Meng, R., Jiang, Z., Liu, Y., Su, M., Yang, X., Fu, Y., Qin, C., Chen, Z., Xu,
  R., Xiong, C., Zhou, Y., Chen, W., and Yavuz, S.
\newblock {VLM2Vec-V2: Advancing Multimodal Embedding for Videos, Images, and
  Visual Documents}.
\newblock \url{https://github.com/TIGER-AI-Lab/VLM2Vec}, 2025.

\bibitem[Muennighoff et~al.(2023)Muennighoff, Tazi, Magne, and Reimers]{mteb}
Muennighoff, N., Tazi, N., Magne, L., and Reimers, N.
\newblock {MTEB}: Massive text embedding benchmark.
\newblock In \emph{Proceedings of the 17th Conference of the European Chapter
  of the Association for Computational Linguistics}, pp.\  2014--2037, 2023.
\newblock \doi{10.18653/v1/2023.eacl-main.148}.
\newblock URL \url{https://aclanthology.org/2023.eacl-main.148}.

\bibitem[Park et~al.(2023)Park, O'Brien, Cai, Morris, Liang, and
  Bernstein]{generative-agents}
Park, J.~S., O'Brien, J., Cai, C.~J., Morris, M.~R., Liang, P., and Bernstein,
  M.~S.
\newblock Generative agents: Interactive simulacra of human behavior.
\newblock In \emph{Proceedings of the 36th annual acm symposium on user
  interface software and technology}, pp.\  1--22, 2023.

\bibitem[Peng et~al.(2023)Peng, Wang, Dong, Hao, Huang, Ma, and Wei]{kosmos-2}
Peng, Z., Wang, W., Dong, L., Hao, Y., Huang, S., Ma, S., and Wei, F.
\newblock Kosmos-2: Grounding multimodal large language models to the world.
\newblock \emph{ArXiv preprint}, abs/2306.14824, 2023.
\newblock URL \url{https://arxiv.org/abs/2306.14824}.

\bibitem[Putta et~al.(2024)Putta, Mills, Garg, Motwani, Finn, Garg, and
  Rafailov]{agent-q}
Putta, P., Mills, E., Garg, N., Motwani, S., Finn, C., Garg, D., and Rafailov,
  R.
\newblock Agent q: Advanced reasoning and learning for autonomous ai agents.
\newblock \emph{ArXiv preprint}, abs/2408.07199, 2024.
\newblock URL \url{https://arxiv.org/abs/2408.07199}.

\bibitem[Qin et~al.(2025)Qin, Ye, Fang, Wang, Liang, Tian, Zhang, Li, Li,
  Huang, et~al.]{ui-tars}
Qin, Y., Ye, Y., Fang, J., Wang, H., Liang, S., Tian, S., Zhang, J., Li, J.,
  Li, Y., Huang, S., et~al.
\newblock Ui-tars: Pioneering automated gui interaction with native agents.
\newblock \emph{ArXiv preprint}, abs/2501.12326, 2025.
\newblock URL \url{https://arxiv.org/abs/2501.12326}.

\bibitem[Radford et~al.(2021)Radford, Kim, Hallacy, Ramesh, Goh, Agarwal,
  Sastry, Askell, Mishkin, Clark, Krueger, and Sutskever]{clip}
Radford, A., Kim, J.~W., Hallacy, C., Ramesh, A., Goh, G., Agarwal, S., Sastry,
  G., Askell, A., Mishkin, P., Clark, J., Krueger, G., and Sutskever, I.
\newblock Learning transferable visual models from natural language
  supervision.
\newblock In \emph{Proceedings of the 38th International Conference on Machine
  Learning, {ICML} 2021, 18-24 July 2021, Virtual Event}, volume 139 of
  \emph{Proceedings of Machine Learning Research}, pp.\  8748--8763, 2021.
\newblock URL \url{http://proceedings.mlr.press/v139/radford21a.html}.

\bibitem[Raposo et~al.(2024)Raposo, Ritter, Richards, Lillicrap, Humphreys, and
  Santoro]{mixture-of-depth}
Raposo, D., Ritter, S., Richards, B., Lillicrap, T., Humphreys, P.~C., and
  Santoro, A.
\newblock Mixture-of-depths: Dynamically allocating compute in
  transformer-based language models.
\newblock \emph{ArXiv preprint}, abs/2404.02258, 2024.
\newblock URL \url{https://arxiv.org/abs/2404.02258}.

\bibitem[Rawles et~al.(2023)Rawles, Li, Rodriguez, Riva, and
  Lillicrap]{android-in-the-wild}
Rawles, C., Li, A., Rodriguez, D., Riva, O., and Lillicrap, T.~P.
\newblock Androidinthewild: {A} large-scale dataset for android device control.
\newblock In \emph{Advances in Neural Information Processing Systems 36: Annual
  Conference on Neural Information Processing Systems 2023, NeurIPS 2023, New
  Orleans, LA, USA, December 10 - 16, 2023}, 2023.

\bibitem[Rawles et~al.(2024)Rawles, Clinckemaillie, Chang, Waltz, Lau, Fair,
  Li, Bishop, Li, Campbell-Ajala, et~al.]{android-world}
Rawles, C., Clinckemaillie, S., Chang, Y., Waltz, J., Lau, G., Fair, M., Li,
  A., Bishop, W., Li, W., Campbell-Ajala, F., et~al.
\newblock Androidworld: A dynamic benchmarking environment for autonomous
  agents.
\newblock \emph{ArXiv preprint}, abs/2405.14573, 2024.
\newblock URL \url{https://arxiv.org/abs/2405.14573}.

\bibitem[Shridhar et~al.(2021)Shridhar, Yuan, C{\^{o}}t{\'{e}}, Bisk,
  Trischler, and Hausknecht]{alfworld}
Shridhar, M., Yuan, X., C{\^{o}}t{\'{e}}, M., Bisk, Y., Trischler, A., and
  Hausknecht, M.~J.
\newblock Alfworld: Aligning text and embodied environments for interactive
  learning.
\newblock In \emph{9th International Conference on Learning Representations,
  {ICLR} 2021, Virtual Event, Austria, May 3-7, 2021}, 2021.
\newblock URL \url{https://openreview.net/forum?id=0IOX0YcCdTn}.

\bibitem[Sridhar et~al.(2024)Sridhar, Dutta, Jayaraman, and Lee]{regent}
Sridhar, K., Dutta, S., Jayaraman, D., and Lee, I.
\newblock Regent: A retrieval-augmented generalist agent that can act
  in-context in new environments.
\newblock \emph{ArXiv preprint}, abs/2412.04759, 2024.
\newblock URL \url{https://arxiv.org/abs/2412.04759}.

\bibitem[Su et~al.(2024)Su, Yang, Yao, and Yu]{language-agent-survey}
Su, Y., Yang, D., Yao, S., and Yu, T.
\newblock Language agents: Foundations, prospects, and risks.
\newblock In \emph{Proceedings of the 2024 Conference on Empirical Methods in
  Natural Language Processing: Tutorial Abstracts}, pp.\  17--24, 2024.
\newblock \doi{10.18653/v1/2024.emnlp-tutorials.3}.

\bibitem[Tang et~al.(2019)Tang, Ding, Rao, Zheng, Zhang, Zhao, Lu, and
  Zhou]{coin}
Tang, Y., Ding, D., Rao, Y., Zheng, Y., Zhang, D., Zhao, L., Lu, J., and Zhou,
  J.
\newblock {COIN:} {A} large-scale dataset for comprehensive instructional video
  analysis.
\newblock In \emph{{IEEE} Conference on Computer Vision and Pattern
  Recognition, {CVPR} 2019, Long Beach, CA, USA, June 16-20, 2019}, pp.\
  1207--1216, 2019.
\newblock \doi{10.1109/CVPR.2019.00130}.

\bibitem[Wang et~al.(2024{\natexlab{a}})Wang, Bai, Tan, Wang, Fan, Bai, Chen,
  Liu, Wang, Ge, et~al.]{qwen2-vl}
Wang, P., Bai, S., Tan, S., Wang, S., Fan, Z., Bai, J., Chen, K., Liu, X.,
  Wang, J., Ge, W., et~al.
\newblock Qwen2-vl: Enhancing vision-language model's perception of the world
  at any resolution.
\newblock \emph{ArXiv preprint}, abs/2409.12191, 2024{\natexlab{a}}.
\newblock URL \url{https://arxiv.org/abs/2409.12191}.

\bibitem[Wang et~al.(2024{\natexlab{b}})Wang, Li, Li, Yu, He, Chen, Pei, Zheng,
  Wang, Shi, et~al.]{internvideo2}
Wang, Y., Li, K., Li, X., Yu, J., He, Y., Chen, G., Pei, B., Zheng, R., Wang,
  Z., Shi, Y., et~al.
\newblock Internvideo2: Scaling foundation models for multimodal video
  understanding.
\newblock In \emph{European Conference on Computer Vision}, pp.\  396--416.
  Springer, 2024{\natexlab{b}}.

\bibitem[Wang et~al.(2024{\natexlab{c}})Wang, Mao, Fried, and Neubig]{awm}
Wang, Z.~Z., Mao, J., Fried, D., and Neubig, G.
\newblock Agent workflow memory.
\newblock \emph{ArXiv preprint}, abs/2409.07429, 2024{\natexlab{c}}.
\newblock URL \url{https://arxiv.org/abs/2409.07429}.

\bibitem[Wei et~al.(2024)Wei, Chen, Chen, Hu, Zhang, Fu, Ritter, and
  Chen]{uniir}
Wei, C., Chen, Y., Chen, H., Hu, H., Zhang, G., Fu, J., Ritter, A., and Chen,
  W.
\newblock Uniir: Training and benchmarking universal multimodal information
  retrievers.
\newblock In \emph{European Conference on Computer Vision}, pp.\  387--404.
  Springer, 2024.

\bibitem[Xie et~al.(2024{\natexlab{a}})Xie, Zhang, Chen, Zhu, Lou, Tian, Xiao,
  and Su]{travelplanner}
Xie, J., Zhang, K., Chen, J., Zhu, T., Lou, R., Tian, Y., Xiao, Y., and Su, Y.
\newblock Travelplanner: {A} benchmark for real-world planning with language
  agents.
\newblock In \emph{Forty-first International Conference on Machine Learning,
  {ICML} 2024, Vienna, Austria, July 21-27, 2024}, 2024{\natexlab{a}}.
\newblock URL \url{https://openreview.net/forum?id=l5XQzNkAOe}.

\bibitem[Xie et~al.(2024{\natexlab{b}})Xie, Zhang, Chen, Li, Zhao, Cao, Hua,
  Cheng, Shin, Lei, Liu, Xu, Zhou, Savarese, Xiong, Zhong, and Yu]{osworld}
Xie, T., Zhang, D., Chen, J., Li, X., Zhao, S., Cao, R., Hua, T.~J., Cheng, Z.,
  Shin, D., Lei, F., Liu, Y., Xu, Y., Zhou, S., Savarese, S., Xiong, C., Zhong,
  V., and Yu, T.
\newblock Osworld: Benchmarking multimodal agents for open-ended tasks in real
  computer environments.
\newblock In \emph{Advances in Neural Information Processing Systems 38: Annual
  Conference on Neural Information Processing Systems 2024, NeurIPS 2024,
  Vancouver, BC, Canada, December 10 - 15, 2024}, 2024{\natexlab{b}}.

\bibitem[Xu et~al.(2024)Xu, Wang, Wang, Lu, Xie, Saha, Sahoo, Yu, and
  Xiong]{aguvis}
Xu, Y., Wang, Z., Wang, J., Lu, D., Xie, T., Saha, A., Sahoo, D., Yu, T., and
  Xiong, C.
\newblock Aguvis: Unified pure vision agents for autonomous gui interaction.
\newblock \emph{ArXiv preprint}, abs/2412.04454, 2024.
\newblock URL \url{https://arxiv.org/abs/2412.04454}.

\bibitem[Yao et~al.(2022)Yao, Chen, Yang, and Narasimhan]{webshop}
Yao, S., Chen, H., Yang, J., and Narasimhan, K.
\newblock Webshop: Towards scalable real-world web interaction with grounded
  language agents.
\newblock In \emph{Advances in Neural Information Processing Systems 35: Annual
  Conference on Neural Information Processing Systems 2022, NeurIPS 2022, New
  Orleans, LA, USA, November 28 - December 9, 2022}, 2022.

\bibitem[Yao et~al.(2023)Yao, Zhao, Yu, Du, Shafran, Narasimhan, and
  Cao]{react}
Yao, S., Zhao, J., Yu, D., Du, N., Shafran, I., Narasimhan, K.~R., and Cao, Y.
\newblock React: Synergizing reasoning and acting in language models.
\newblock In \emph{The Eleventh International Conference on Learning
  Representations, {ICLR} 2023, Kigali, Rwanda, May 1-5, 2023}, 2023.
\newblock URL \url{https://openreview.net/pdf?id=WE\_vluYUL-X}.

\bibitem[Yue et~al.(2024)Yue, Xu, Karlsson, and Lu]{embodied-mllm-retriever}
Yue, J., Xu, X., Karlsson, B.~F., and Lu, Z.
\newblock Mllm as retriever: Interactively learning multimodal retrieval for
  embodied agents.
\newblock \emph{ArXiv preprint}, abs/2410.03450, 2024.
\newblock URL \url{https://arxiv.org/abs/2410.03450}.

\bibitem[Zhang et~al.(2025)Zhang, Shang, Gao, Zhang, Xie, Ma, Yuan, Wu, Zhu,
  and Li]{tongui}
Zhang, B., Shang, Z., Gao, Z., Zhang, W., Xie, R., Ma, X., Yuan, T., Wu, X.,
  Zhu, S.-C., and Li, Q.
\newblock Tongui: Building generalized gui agents by learning from multimodal
  web tutorials.
\newblock \emph{ArXiv preprint}, abs/2504.12679, 2025.
\newblock URL \url{https://arxiv.org/abs/2504.12679}.

\bibitem[Zhang et~al.(2024{\natexlab{a}})Zhang, He, Qian, Li, Li, Qin, Kang,
  Ma, Liu, Lin, et~al.]{gui-agent-survey}
Zhang, C., He, S., Qian, J., Li, B., Li, L., Qin, S., Kang, Y., Ma, M., Liu,
  G., Lin, Q., et~al.
\newblock Large language model-brained gui agents: A survey.
\newblock \emph{ArXiv preprint}, abs/2411.18279, 2024{\natexlab{a}}.
\newblock URL \url{https://arxiv.org/abs/2411.18279}.

\bibitem[Zhang et~al.(2024{\natexlab{b}})Zhang, Luan, Hu, Lee, Qiao, Chen, Su,
  and Chang]{magiclens}
Zhang, K., Luan, Y., Hu, H., Lee, K., Qiao, S., Chen, W., Su, Y., and Chang, M.
\newblock Magiclens: Self-supervised image retrieval with open-ended
  instructions.
\newblock In \emph{Forty-first International Conference on Machine Learning,
  {ICML} 2024, Vienna, Austria, July 21-27, 2024}, 2024{\natexlab{b}}.
\newblock URL \url{https://openreview.net/forum?id=Zc22RDtsvP}.

\bibitem[Zhang et~al.(2024{\natexlab{c}})Zhang, Deng, Ren, Ng, and
  Chua]{ask-before-plan}
Zhang, X., Deng, Y., Ren, Z., Ng, S.-K., and Chua, T.-S.
\newblock Ask-before-plan: Proactive language agents for real-world planning.
\newblock \emph{ArXiv preprint}, abs/2406.12639, 2024{\natexlab{c}}.
\newblock URL \url{https://arxiv.org/abs/2406.12639}.

\bibitem[Zhang et~al.(2024{\natexlab{d}})Zhang, Zhang, Xie, Li, Dai, Long, Xie,
  Zhang, Li, and Zhang]{gme}
Zhang, X., Zhang, Y., Xie, W., Li, M., Dai, Z., Long, D., Xie, P., Zhang, M.,
  Li, W., and Zhang, M.
\newblock Gme: Improving universal multimodal retrieval by multimodal llms.
\newblock \emph{ArXiv preprint}, abs/2412.16855, 2024{\natexlab{d}}.
\newblock URL \url{https://arxiv.org/abs/2412.16855}.

\bibitem[Zhao et~al.(2024)Zhao, Huang, Xu, Lin, Liu, and Huang]{expel}
Zhao, A., Huang, D., Xu, Q., Lin, M., Liu, Y., and Huang, G.
\newblock Expel: {LLM} agents are experiential learners.
\newblock In \emph{Thirty-Eighth {AAAI} Conference on Artificial Intelligence,
  {AAAI} 2024, Thirty-Sixth Conference on Innovative Applications of Artificial
  Intelligence, {IAAI} 2024, Fourteenth Symposium on Educational Advances in
  Artificial Intelligence, {EAAI} 2014, February 20-27, 2024, Vancouver,
  Canada}, pp.\  19632--19642, 2024.
\newblock \doi{10.1609/AAAI.V38I17.29936}.
\newblock URL \url{https://doi.org/10.1609/aaai.v38i17.29936}.

\bibitem[Zheng et~al.(2025)Zheng, Fatemi, Jin, Wang, Gandhi, Song, Gu,
  Srinivasa, Liu, Neubig, et~al.]{skillweaver}
Zheng, B., Fatemi, M.~Y., Jin, X., Wang, Z.~Z., Gandhi, A., Song, Y., Gu, Y.,
  Srinivasa, J., Liu, G., Neubig, G., et~al.
\newblock Skillweaver: Web agents can self-improve by discovering and honing
  skills.
\newblock \emph{ArXiv preprint}, abs/2504.07079, 2025.
\newblock URL \url{https://arxiv.org/abs/2504.07079}.

\bibitem[Zhou et~al.(2022)Zhou, Zhang, Yang, Lyu, Yin, Callison-Burch, and
  Neubig]{show-me-more-details}
Zhou, S., Zhang, L., Yang, Y., Lyu, Q., Yin, P., Callison-Burch, C., and
  Neubig, G.
\newblock Show me more details: Discovering hierarchies of procedures from
  semi-structured web data.
\newblock In \emph{Proceedings of the 60th Annual Meeting of the Association
  for Computational Linguistics (Volume 1: Long Papers)}, pp.\  2998--3012,
  2022.
\newblock \doi{10.18653/v1/2022.acl-long.214}.
\newblock URL \url{https://aclanthology.org/2022.acl-long.214}.

\bibitem[Zhou et~al.(2024)Zhou, Xu, Zhu, Zhou, Lo, Sridhar, Cheng, Ou, Bisk,
  Fried, Alon, and Neubig]{webarena}
Zhou, S., Xu, F.~F., Zhu, H., Zhou, X., Lo, R., Sridhar, A., Cheng, X., Ou, T.,
  Bisk, Y., Fried, D., Alon, U., and Neubig, G.
\newblock Webarena: {A} realistic web environment for building autonomous
  agents.
\newblock In \emph{The Twelfth International Conference on Learning
  Representations, {ICLR} 2024, Vienna, Austria, May 7-11, 2024}, 2024.
\newblock URL \url{https://openreview.net/forum?id=oKn9c6ytLx}.

\end{thebibliography}
    \bibliographystyle{icml2025}

    \newpage
    \appendix
    \section{Input Sequence Formation}
\label{app:model_input_design}
As described in Subsections~\ref{subsec:uatd} and~\ref{subsec:problem_definition}, we represent a state $s_i$ using a screenshot, and each action $a_i$ is expressed as a JSON object with three keys: \textit{operation}, \textit{value}, and \textit{target}. A trajectory subsequence $\tau_{i:j}$ or a full trajectory $\tau_{1:n}$ (or simply $\tau$) is composed of state-action pairs. To standardize the representation across different environments, we prepend the action space definition to $\tau_{i:j}$ and $\tau_{1:n}$, resulting in $u_{i:j}$ and $u_{1:n}$. For the target side of retrieval, $\mathbf{v}$ can be either a single state $s_i$ or a trajectory sequence with action definitions, denoted as $u_{i:j}$. For the query side $\mathbf{k}$, we prepend the retrieval query $\tilde{q}$, which includes a task-specific instruction and trajectory description to $s_i$ or  $u_{i:j}$. Table~\ref{tab:input_repr} displays concrete examples, where \texttt{[image]} denotes image tokens.

\begin{table}[h]
\scriptsize
\centering
\vspace{-0.3cm}
\setlength{\abovecaptionskip}{0pt}   
\setlength{\belowcaptionskip}{0pt}
\caption{Examples of model input representations for a state ($s_i$), an action ($a_i$), a trajectory sequence or subsequence with action definitions ($u_{i:j}$), and an augmented retrieval query ($\tilde{q}$). The retrieval key $\mathbf{k}$ consists of $\tilde{q}$ combined with either $s_i$ or $u_{i:j}$, while the retrieval value $\mathbf{v}$ is composed of either $s_i$ or $u_{i:j}$.}

\begin{tabularx}{\columnwidth}{lX}
    \toprule
    \multicolumn{1}{c}{\textbf{Variable}} 
      & \multicolumn{1}{c}{\textbf{Representation Instance}} \\
    \midrule
    $s_i$
      & Observation: \texttt{[image]} \\
    \midrule
    $a_i$
      & Action $i$: \{"operation": "type", "value": "Elon Musk", "target": \{"x": 0.5704, "y": 0.2142, "width": 0.3678, "height": 0.0663\}\}\\
    \midrule
    $u_{i,j}$ & Action Space:\newline 1. click: Simulates a mouse click on the target bounding box.\newline 2. type: Types the value (str) into the target bounding box.\newline Positions are represented in relative coordinates within the range [0,1] on the observation screenshot.\newline Observation 1: \texttt{[image]}\newline Action 1: \{"operation": "click", "value": null, "target": \{"x": 0.0021, "y": 0.1424, "width": 0.0243, "height": 0.0519\}\}\newline Observation 2: \texttt{[image]}\newline Action 2: \{"operation": "type", "value": "", "target": \{"x": 0.8343, "y": 0.5659, "width": 0.1034, "height": 0.2496\}\}\\
    \midrule
    $\tilde{q}$ & Apply the request "Creating a template on Trello in a new tab." to the previous web navigation steps to derive the next trajectory.\\
    \bottomrule
\end{tabularx}
\label{tab:input_repr}
\vspace{-0.3cm}
\end{table}

\section{Subtask Performance} 
\label{app:subtask_performance}
In this section, we present the evaluation of all baseline methods described in Subsection~\ref{subsec:experimental_setups}, along with our proposed GAE-Retriever, on 12 retrieval subtasks spanning all data sources. Results are given in terms of Recall@1/5/10. Table~\ref{tab:subtask_recall_ind} provides the outcomes under the in-domain setting, while Table~\ref{tab:subtask_recall_ood} covers the out-of-domain evaluation.

\section{Prompt List} 
\label{app:prompt_list}
This section demonstrates the prompts used in the paper.
Table~\ref{prompt:data_annotation} includes prompts for data annotation such as state description generation, HTML rendering, and silver data generation.
Snippet~\ref{prompt:action_space} specifies the action space definitions used for Mind2Web, WebLINX, WebArena, GUIAct, and AutoWebGLM.
Snippets~\ref{prompt:instruction_template} and~\ref{prompt:instruction_template_cont} show the instruction templates used to augment retrieval queries, where \textit{description} refers to the natural language summary of a trajectory.

\onecolumn

\begin{table}[ht]
  \centering
  \small
  \setlength{\abovecaptionskip}{5pt}
  \setlength{\belowcaptionskip}{0pt}
  \caption{Recall@1/5/10 for all methods on each subtask in the in-domain setting.}
  \begin{adjustbox}{max width=\columnwidth}
    \begin{tabular}{@{}l *{12}{c}@{}}
      \toprule
      \textbf{Source}
        & \textbf{$(q,\tau_{1:i})\to\tau_{i+1:n}$}
        & \textbf{$(q,\tau_{i+1:n})\to\tau_{1:i}$}
        & \textbf{$(q,\tau_{1:i})\to s_{i+1}$}
        & \textbf{$(q,\tau_{i+1:n})\to s_i$}
        & \textbf{$q\to\tau_{\equiv}$}
        & \textbf{$q\to\tau_{\sim}$}
        & \textbf{$(q,s_i)\to s_{i+1}$}
        & \textbf{$(q,s_{i+1})\to s_i$}
        & \textbf{$(q,s_i)\to\tau_{i+1:n}$}
        & \textbf{$(q,s_{i+1})\to\tau_{1:i}$}
        & \textbf{$q\to s_i$}
        & \textbf{$q\to s_n$} \\
      \midrule

      \rowcolor{blue!30}\multicolumn{13}{c}{\bfseries Qwen2-VL-2B} \\
      \midrule
      Mind2Web & \trip{0.0}{0.0}{1.5} & \trip{0.0}{1.0}{2.0} & \trip{0.1}{0.3}{0.6} & \trip{0.0}{0.1}{0.7} & \trip{0.0}{0.9}{1.8} & \trip{0.0}{1.5}{2.0} & \trip{3.2}{54.9}{70.8} & \trip{2.6}{55.0}{71.2} & \trip{0.0}{2.0}{2.5} & \trip{0.0}{0.0}{0.0} & \trip{0.0}{0.3}{0.6} & \trip{0.0}{0.7}{0.7} \\
      AutoWebGLM & \trip{0.0}{0.0}{4.0} & \trip{0.0}{0.0}{0.0} & \trip{2.0}{2.0}{4.0} & \trip{0.0}{0.0}{0.0} & \trip{7.1}{17.9}{35.7} & \trip{8.9}{16.1}{39.3} & \trip{0.0}{18.0}{26.0} & \trip{0.0}{18.0}{32.0} & \trip{2.0}{6.0}{8.0} & \trip{0.0}{0.0}{0.0} & \trip{0.0}{0.0}{0.0} & \trip{3.3}{3.3}{3.3} \\
      WebArena & \trip{0.0}{0.0}{0.0} & \trip{0.0}{0.0}{0.0} & \trip{0.0}{2.4}{3.6} & \trip{0.0}{3.6}{6.0} & \trip{0.0}{10.8}{21.6} & \trip{0.0}{9.3}{20.0} & \trip{3.1}{39.6}{51.0} & \trip{10.4}{47.9}{54.2} & \trip{0.0}{0.0}{2.4} & \trip{0.0}{0.0}{0.0} & \trip{0.0}{0.0}{1.9} & \trip{0.0}{0.0}{0.0} \\
      WebLINX & \trip{0.0}{0.5}{0.5} & \trip{0.0}{0.5}{1.0} & \trip{0.3}{0.8}{1.3} & \trip{0.0}{0.0}{0.0} & \trip{0.0}{4.3}{10.9} & \trip{1.1}{5.4}{15.1} & \trip{1.9}{35.8}{48.1} & \trip{17.5}{48.2}{58.3} & \trip{0.0}{0.0}{0.5} & \trip{0.0}{1.0}{1.0} & \trip{0.0}{0.2}{0.5} & \trip{0.0}{0.0}{0.0} \\
      GUIAct & \trip{0.0}{0.0}{0.0} & \trip{0.0}{0.0}{0.0} & \trip{0.1}{0.1}{0.3} & \trip{0.0}{0.2}{0.4} & \trip{1.0}{4.0}{4.5} & \trip{0.0}{0.0}{0.5} & \trip{0.6}{18.9}{23.8} & \trip{23.4}{35.9}{39.1} & \trip{0.0}{0.0}{0.0} & \trip{0.0}{0.0}{0.0} & \trip{0.0}{0.1}{0.3} & \trip{0.0}{0.2}{0.2} \\
      \midrule

\rowcolor{blue!30}\multicolumn{13}{c}{\bfseries Qwen2.5-VL-3B} \\
      \midrule
      Mind2Web & \trip{0.0}{0.0}{0.5} & \trip{0.0}{0.0}{0.0} & \trip{0.0}{0.7}{0.7} & \trip{0.0}{0.1}{0.3} & \trip{0.9}{2.7}{12.4} & \trip{1.5}{4.5}{10.0} & \trip{3.7}{30.7}{39.2} & \trip{4.1}{30.3}{37.7} & \trip{0.0}{0.0}{0.0} & \trip{0.0}{0.0}{0.0} & \trip{0.2}{0.5}{0.6} & \trip{0.0}{0.7}{0.7} \\
      AutoWebGLM & \trip{0.0}{0.0}{0.0} & \trip{0.0}{0.0}{0.0} & \trip{0.0}{2.0}{4.0} & \trip{0.0}{2.0}{4.0} & \trip{3.6}{14.3}{21.4} & \trip{1.8}{14.3}{19.6} & \trip{0.0}{4.0}{8.0} & \trip{0.0}{0.0}{2.0} & \trip{0.0}{0.0}{0.0} & \trip{2.0}{2.0}{2.0} & \trip{0.0}{1.8}{3.6} & \trip{0.0}{3.3}{6.7} \\
      WebArena & \trip{0.0}{0.0}{0.0} & \trip{0.0}{0.0}{0.0} & \trip{0.0}{0.0}{2.4} & \trip{0.0}{0.0}{1.2} & \trip{2.7}{8.1}{24.3} & \trip{2.7}{8.0}{20.0} & \trip{2.1}{34.4}{42.7} & \trip{4.2}{37.5}{44.8} & \trip{0.0}{1.2}{1.2} & \trip{0.0}{0.0}{0.0} & \trip{0.0}{1.0}{1.0} & \trip{0.0}{2.2}{4.4} \\
      WebLINX & \trip{0.0}{1.0}{1.0} & \trip{0.5}{0.5}{1.0} & \trip{0.0}{0.3}{0.3} & \trip{0.0}{0.3}{0.3} & \trip{2.2}{8.7}{23.9} & \trip{2.2}{11.8}{21.5} & \trip{5.1}{34.7}{41.7} & \trip{16.6}{44.7}{53.2} & \trip{0.0}{0.0}{0.0} & \trip{0.0}{0.0}{0.5} & \trip{0.0}{0.2}{0.3} & \trip{0.0}{2.0}{2.0} \\
      GUIAct & \trip{0.0}{0.0}{0.0} & \trip{0.0}{0.0}{0.0} & \trip{0.0}{0.1}{0.1} & \trip{0.0}{0.1}{0.2} & \trip{1.0}{1.0}{2.5} & \trip{0.0}{2.5}{5.0} & \trip{1.9}{18.7}{23.4} & \trip{20.6}{33.0}{36.9} & \trip{0.0}{0.0}{0.0} & \trip{0.0}{0.0}{0.0} & \trip{0.0}{0.0}{0.0} & \trip{0.0}{0.0}{0.0} \\
      \midrule

\rowcolor{blue!30}\multicolumn{13}{c}{\bfseries LamRA-Ret} \\
      \midrule
      Mind2Web & \trip{0.0}{0.0}{0.0} & \trip{0.0}{0.0}{0.0} & \trip{0.0}{0.3}{0.6} & \trip{0.0}{0.7}{1.2} & \trip{6.2}{13.3}{19.5} & \trip{0.5}{9.0}{16.0} & \trip{2.6}{54.5}{69.3} & \trip{3.3}{54.6}{70.1} & \trip{0.0}{0.0}{0.5} & \trip{0.0}{0.0}{0.0} & \trip{1.0}{5.0}{7.7} & \trip{0.7}{2.2}{4.5} \\
      AutoWebGLM & \trip{0.0}{0.0}{0.0} & \trip{0.0}{0.0}{0.0} & \trip{0.0}{10.0}{16.0} & \trip{4.0}{6.0}{16.0} & \trip{25.0}{39.3}{53.6} & \trip{16.1}{35.7}{50.0} & \trip{0.0}{18.0}{28.0} & \trip{0.0}{24.0}{36.0} & \trip{2.0}{6.0}{16.0} & \trip{0.0}{0.0}{0.0} & \trip{0.0}{7.1}{17.9} & \trip{0.0}{6.7}{16.7} \\
      WebArena & \trip{0.0}{0.0}{1.4} & \trip{0.0}{0.0}{0.0} & \trip{0.0}{2.4}{4.8} & \trip{0.0}{2.4}{4.8} & \trip{13.5}{32.4}{48.6} & \trip{6.7}{22.7}{33.3} & \trip{0.0}{34.4}{49.0} & \trip{9.4}{40.6}{51.0} & \trip{0.0}{0.0}{0.0} & \trip{0.0}{1.2}{1.2} & \trip{0.0}{0.0}{1.9} & \trip{0.0}{0.0}{4.4} \\
      WebLINX & \trip{0.0}{0.0}{1.0} & \trip{0.0}{0.0}{0.5} & \trip{0.3}{0.5}{1.1} & \trip{0.0}{0.0}{0.3} & \trip{4.3}{21.7}{32.6} & \trip{2.2}{5.4}{23.7} & \trip{0.5}{38.9}{48.9} & \trip{17.8}{50.3}{58.3} & \trip{0.0}{0.0}{0.0} & \trip{0.0}{0.0}{0.0} & \trip{1.7}{7.0}{12.0} & \trip{2.0}{6.0}{14.0} \\
      GUIAct & \trip{0.0}{0.0}{0.0} & \trip{0.0}{0.0}{0.0} & \trip{0.1}{0.1}{0.3} & \trip{0.1}{0.4}{0.7} & \trip{21.0}{46.5}{56.5} & \trip{9.0}{24.0}{32.0} & \trip{0.7}{18.7}{23.6} & \trip{23.1}{35.1}{38.2} & \trip{0.5}{0.5}{0.5} & \trip{0.0}{0.0}{0.5} & \trip{0.3}{0.8}{1.5} & \trip{0.2}{0.6}{0.8} \\
      \midrule

\rowcolor{blue!30}\multicolumn{13}{c}{\bfseries ColQwen2-v1.0} \\
      \midrule
      Mind2Web & \trip{0.0}{0.0}{0.0} & \trip{0.0}{0.0}{0.0} & \trip{1.6}{8.9}{14.2} & \trip{1.9}{7.7}{13.5} & \trip{9.7}{23.0}{36.3} & \trip{6.0}{17.0}{28.0} & \trip{2.4}{56.2}{73.6} & \trip{3.2}{56.2}{74.1} & \trip{0.0}{1.5}{3.0} & \trip{1.0}{5.0}{8.5} & \trip{3.2}{8.0}{13.8} & \trip{1.5}{9.0}{11.2} \\
      AutoWebGLM & \trip{0.0}{0.0}{0.0} & \trip{0.0}{0.0}{2.0} & \trip{6.0}{22.0}{28.0} & \trip{0.0}{8.0}{22.0} & \trip{21.4}{35.7}{53.6} & \trip{21.4}{30.4}{51.8} & \trip{0.0}{24.0}{42.0} & \trip{0.0}{30.0}{50.0} & \trip{0.0}{0.0}{0.0} & \trip{0.0}{6.0}{12.0} & \trip{5.4}{8.9}{16.1} & \trip{16.7}{26.7}{26.7} \\
      WebArena & \trip{0.0}{0.0}{0.0} & \trip{0.0}{1.4}{1.4} & \trip{2.4}{3.6}{10.7} & \trip{1.2}{11.9}{16.7} & \trip{8.1}{27.0}{54.1} & \trip{2.7}{25.3}{44.0} & \trip{3.1}{42.7}{52.1} & \trip{11.5}{43.8}{52.1} & \trip{0.0}{0.0}{0.0} & \trip{1.2}{4.8}{7.1} & \trip{1.9}{3.9}{9.7} & \trip{0.0}{8.9}{13.3} \\
      WebLINX & \trip{0.0}{0.0}{1.0} & \trip{0.0}{0.0}{1.5} & \trip{1.6}{6.9}{11.5} & \trip{1.3}{10.1}{15.7} & \trip{21.7}{58.7}{67.4} & \trip{9.7}{34.4}{45.2} & \trip{1.1}{39.3}{51.6} & \trip{17.2}{48.6}{59.9} & \trip{0.5}{0.5}{2.5} & \trip{1.5}{3.0}{3.5} & \trip{0.3}{4.4}{7.5} & \trip{4.0}{10.0}{12.0} \\
      GUIAct & \trip{0.0}{0.0}{0.5} & \trip{0.0}{0.0}{0.0} & \trip{0.4}{1.6}{2.6} & \trip{1.1}{4.7}{7.6} & \trip{57.0}{81.0}{89.0} & \trip{20.5}{48.5}{63.0} & \trip{0.7}{24.1}{30.1} & \trip{23.3}{38.9}{42.9} & \trip{0.5}{4.0}{8.5} & \trip{1.0}{1.5}{2.0} & \trip{0.8}{1.9}{3.2} & \trip{0.4}{1.7}{2.3} \\
      \midrule

\rowcolor{blue!30}\multicolumn{13}{c}{\bfseries GME-Qwen2VL-2B} \\
      \midrule
      Mind2Web & \trip{0.0}{0.0}{1.0} & \trip{0.0}{0.0}{0.0} & \trip{2.0}{8.7}{16.7} & \trip{1.5}{8.4}{15.8} & \trip{13.3}{39.8}{55.8} & \trip{10.5}{24.0}{36.5} & \trip{2.0}{57.5}{74.7} & \trip{2.8}{55.8}{74.1} & \trip{0.5}{2.0}{6.5} & \trip{0.5}{4.0}{8.5} & \trip{2.5}{7.2}{12.8} & \trip{1.5}{6.7}{9.0} \\
      AutoWebGLM & \trip{0.0}{0.0}{0.0} & \trip{0.0}{0.0}{0.0} & \trip{6.0}{16.0}{30.0} & \trip{6.0}{18.0}{28.0} & \trip{71.4}{82.1}{92.9} & \trip{30.4}{60.7}{75.0} & \trip{0.0}{40.0}{60.0} & \trip{0.0}{46.0}{60.0} & \trip{0.0}{0.0}{2.0} & \trip{0.0}{4.0}{6.0} & \trip{5.4}{16.1}{26.8} & \trip{13.3}{33.3}{50.0} \\
      WebArena & \trip{0.0}{0.0}{0.0} & \trip{0.0}{0.0}{0.0} & \trip{1.2}{9.5}{16.7} & \trip{3.6}{16.7}{22.6} & \trip{16.2}{45.9}{59.5} & \trip{10.7}{36.0}{53.3} & \trip{2.1}{40.6}{52.1} & \trip{11.5}{47.9}{53.1} & \trip{1.2}{2.4}{7.1} & \trip{1.2}{2.4}{4.8} & \trip{4.9}{16.5}{18.4} & \trip{0.0}{0.0}{17.8} \\
      WebLINX & \trip{0.0}{0.0}{1.5} & \trip{0.0}{0.5}{1.0} & \trip{2.1}{14.1}{24.3} & \trip{4.3}{13.6}{25.6} & \trip{34.8}{63.0}{73.9} & \trip{11.8}{40.9}{57.0} & \trip{1.1}{40.4}{51.9} & \trip{17.0}{50.3}{62.3} & \trip{0.0}{4.0}{6.5} & \trip{0.0}{4.0}{7.0} & \trip{0.5}{4.2}{7.2} & \trip{2.0}{6.0}{6.0} \\
      GUIAct & \trip{0.0}{0.0}{0.0} & \trip{0.0}{0.0}{0.0} & \trip{0.9}{4.4}{6.9} & \trip{1.7}{8.0}{11.6} & \trip{50.0}{80.5}{93.0} & \trip{10.0}{34.5}{48.0} & \trip{1.1}{24.4}{30.7} & \trip{22.1}{39.4}{43.7} & \trip{0.0}{4.0}{5.0} & \trip{1.5}{4.0}{5.5} & \trip{0.9}{2.8}{4.2} & \trip{0.4}{1.0}{2.7} \\
      \midrule

\rowcolor{blue!30}\multicolumn{13}{c}{\bfseries UniSE-MLLM} \\
      \midrule
      Mind2Web & \trip{0.0}{0.0}{0.5} & \trip{0.0}{0.5}{4.0} & \trip{0.9}{3.1}{6.1} & \trip{0.7}{3.9}{7.6} & \trip{5.3}{9.7}{13.3} & \trip{1.5}{6.0}{11.5} & \trip{5.8}{57.1}{73.4} & \trip{4.5}{55.3}{74.8} & \trip{2.0}{4.0}{5.5} & \trip{0.0}{0.5}{1.0} & \trip{1.7}{6.0}{11.5} & \trip{0.7}{7.5}{9.0} \\
      AutoWebGLM & \trip{0.0}{0.0}{0.0} & \trip{0.0}{0.0}{2.0} & \trip{4.0}{16.0}{20.0} & \trip{0.0}{4.0}{12.0} & \trip{25.0}{32.1}{39.3} & \trip{8.9}{19.6}{37.5} & \trip{0.0}{24.0}{34.0} & \trip{0.0}{30.0}{38.0} & \trip{0.0}{0.0}{4.0} & \trip{0.0}{0.0}{2.0} & \trip{0.0}{3.6}{5.4} & \trip{0.0}{3.3}{6.7} \\
      WebArena & \trip{0.0}{0.0}{0.0} & \trip{0.0}{1.4}{2.7} & \trip{1.2}{4.8}{7.1} & \trip{0.0}{13.1}{15.5} & \trip{8.1}{13.5}{37.8} & \trip{5.3}{14.7}{32.0} & \trip{5.2}{42.7}{53.1} & \trip{9.4}{44.8}{54.2} & \trip{1.2}{4.8}{7.1} & \trip{0.0}{0.0}{0.0} & \trip{1.9}{5.8}{9.7} & \trip{0.0}{6.7}{8.9} \\
      WebLINX & \trip{0.0}{1.0}{2.0} & \trip{0.0}{0.0}{1.5} & \trip{0.5}{3.7}{5.1} & \trip{0.0}{0.8}{1.6} & \trip{2.2}{8.7}{21.7} & \trip{1.1}{7.5}{22.6} & \trip{3.5}{36.9}{48.1} & \trip{17.5}{48.4}{59.4} & \trip{0.0}{1.5}{2.0} & \trip{0.0}{0.5}{0.5} & \trip{0.3}{0.9}{1.4} & \trip{0.0}{2.0}{2.0} \\
      GUIAct & \trip{0.0}{0.0}{0.5} & \trip{0.0}{0.0}{0.0} & \trip{0.4}{0.4}{0.9} & \trip{0.4}{1.3}{1.6} & \trip{3.5}{10.5}{14.0} & \trip{1.0}{2.5}{3.5} & \trip{0.7}{20.2}{26.6} & \trip{23.2}{37.4}{41.0} & \trip{0.0}{0.0}{0.0} & \trip{0.0}{0.0}{0.0} & \trip{0.1}{0.3}{0.6} & \trip{0.0}{0.2}{0.4} \\
      \midrule

\rowcolor{blue!30}\multicolumn{13}{c}{\bfseries VLM2Vec-Qwen2VL-2B} \\
      \midrule
      Mind2Web & \trip{0.0}{0.0}{0.0} & \trip{0.0}{0.5}{0.5} & \trip{5.5}{27.0}{48.0} & \trip{4.5}{31.8}{57.8} & \trip{24.8}{60.2}{76.1} & \trip{10.5}{33.0}{49.0} & \trip{2.4}{62.4}{79.2} & \trip{2.0}{59.6}{80.2} & \trip{0.5}{5.5}{7.5} & \trip{0.5}{4.5}{5.5} & \trip{9.8}{30.2}{46.7} & \trip{10.4}{33.6}{45.5} \\
      AutoWebGLM & \trip{0.0}{0.0}{0.0} & \trip{0.0}{0.0}{4.0} & \trip{8.0}{26.0}{56.0} & \trip{8.0}{38.0}{66.0} & \trip{67.9}{82.1}{89.3} & \trip{32.1}{58.9}{71.4} & \trip{0.0}{44.0}{64.0} & \trip{0.0}{42.0}{56.0} & \trip{2.0}{8.0}{14.0} & \trip{2.0}{2.0}{8.0} & \trip{16.1}{44.6}{57.1} & \trip{13.3}{40.0}{46.7} \\
      WebArena & \trip{0.0}{0.0}{0.0} & \trip{0.0}{0.0}{0.0} & \trip{2.4}{13.1}{21.4} & \trip{4.8}{20.2}{27.4} & \trip{24.3}{59.5}{78.4} & \trip{14.7}{40.0}{57.3} & \trip{4.2}{43.8}{53.1} & \trip{8.3}{46.9}{59.4} & \trip{0.0}{4.8}{9.5} & \trip{0.0}{2.4}{3.6} & \trip{5.8}{23.3}{47.6} & \trip{13.3}{24.4}{28.9} \\
      WebLINX & \trip{0.0}{0.0}{0.5} & \trip{0.0}{0.0}{0.5} & \trip{5.6}{23.7}{39.7} & \trip{6.4}{23.2}{41.9} & \trip{37.0}{56.5}{78.3} & \trip{14.0}{39.8}{51.6} & \trip{1.3}{43.0}{55.4} & \trip{16.9}{51.1}{65.4} & \trip{1.5}{3.5}{8.5} & \trip{0.5}{2.5}{6.5} & \trip{4.9}{20.1}{30.4} & \trip{2.0}{4.0}{6.0} \\
      GUIAct & \trip{0.0}{0.0}{0.0} & \trip{0.0}{0.0}{0.0} & \trip{0.9}{5.2}{8.0} & \trip{3.9}{14.0}{19.3} & \trip{67.0}{91.0}{95.0} & \trip{15.0}{35.5}{55.0} & \trip{0.3}{28.0}{35.8} & \trip{24.0}{42.3}{47.2} & \trip{7.5}{16.0}{23.5} & \trip{0.0}{1.5}{2.0} & \trip{3.6}{11.1}{16.0} & \trip{0.4}{1.2}{2.1} \\
      \midrule

\rowcolor{blue!30}\multicolumn{13}{c}{\bfseries VLM2Vec-V2.2} \\
      \midrule
      Mind2Web & \trip{0.0}{0.0}{0.0} & \trip{0.0}{0.0}{0.0} & \trip{7.6}{46.8}{70.9} & \trip{4.1}{40.8}{74.7} & \trip{66.4}{90.3}{95.6} & \trip{28.0}{60.0}{76.5} & \trip{5.1}{59.4}{78.1} & \trip{3.8}{58.3}{79.2} & \trip{1.5}{6.0}{10.0} & \trip{1.5}{4.5}{8.0} & \trip{11.8}{37.4}{55.6} & \trip{17.9}{43.3}{61.2} \\
      AutoWebGLM & \trip{0.0}{0.0}{2.0} & \trip{0.0}{0.0}{10.0} & \trip{6.0}{44.0}{80.0} & \trip{4.0}{32.0}{72.0} & \trip{85.7}{100.0}{100.0} & \trip{48.2}{73.2}{83.9} & \trip{0.0}{58.0}{76.0} & \trip{0.0}{58.0}{70.0} & \trip{0.0}{12.0}{16.0} & \trip{2.0}{2.0}{4.0} & \trip{16.1}{62.5}{85.7} & \trip{16.7}{40.0}{66.7} \\
      WebArena & \trip{0.0}{0.0}{0.0} & \trip{0.0}{0.0}{0.0} & \trip{3.6}{21.4}{26.2} & \trip{1.2}{27.4}{44.0} & \trip{45.9}{78.4}{86.5} & \trip{32.0}{61.3}{74.7} & \trip{2.1}{44.8}{52.1} & \trip{10.4}{51.0}{59.4} & \trip{0.0}{4.8}{9.5} & \trip{0.0}{1.2}{4.8} & \trip{7.8}{31.1}{44.7} & \trip{8.9}{35.6}{51.1} \\
      WebLINX & \trip{0.0}{0.0}{0.5} & \trip{0.0}{0.0}{0.5} & \trip{8.5}{44.8}{64.0} & \trip{6.4}{37.3}{62.4} & \trip{78.3}{91.3}{93.5} & \trip{37.6}{68.8}{76.3} & \trip{3.0}{45.4}{59.2} & \trip{18.8}{53.5}{66.7} & \trip{0.0}{4.0}{10.5} & \trip{0.0}{3.0}{6.5} & \trip{14.3}{43.5}{61.1} & \trip{0.0}{4.0}{14.0} \\
      GUIAct & \trip{0.0}{0.0}{0.0} & \trip{0.0}{0.0}{0.0} & \trip{5.0}{21.9}{31.6} & \trip{9.4}{34.0}{45.1} & \trip{96.0}{100.0}{100.0} & \trip{33.5}{67.0}{83.0} & \trip{1.6}{33.2}{41.8} & \trip{23.5}{46.2}{52.2} & \trip{4.5}{11.5}{15.0} & \trip{0.0}{2.5}{4.5} & \trip{19.9}{41.5}{50.0} & \trip{1.6}{6.0}{10.3} \\
      \midrule

\rowcolor{blue!30}\multicolumn{13}{c}{\bfseries UGround-V1-2B} \\
      \midrule
      Mind2Web & \trip{0.0}{0.0}{0.0} & \trip{0.0}{1.0}{3.0} & \trip{0.0}{0.1}{0.1} & \trip{0.0}{0.0}{0.1} & \trip{0.0}{0.0}{0.0} & \trip{0.0}{0.0}{0.0} & \trip{3.4}{51.3}{67.5} & \trip{3.6}{50.5}{67.4} & \trip{0.0}{0.0}{0.0} & \trip{0.5}{0.5}{0.5} & \trip{0.0}{0.1}{0.2} & \trip{0.0}{0.0}{0.7} \\
      AutoWebGLM & \trip{0.0}{0.0}{0.0} & \trip{0.0}{0.0}{0.0} & \trip{0.0}{0.0}{4.0} & \trip{0.0}{0.0}{2.0} & \trip{3.6}{14.3}{25.0} & \trip{3.6}{14.3}{19.6} & \trip{0.0}{14.0}{22.0} & \trip{0.0}{20.0}{22.0} & \trip{0.0}{0.0}{0.0} & \trip{0.0}{2.0}{4.0} & \trip{0.0}{0.0}{1.8} & \trip{0.0}{3.3}{3.3} \\
      WebArena & \trip{0.0}{0.0}{0.0} & \trip{0.0}{0.0}{0.0} & \trip{0.0}{0.0}{3.6} & \trip{0.0}{3.6}{4.8} & \trip{0.0}{0.0}{2.7} & \trip{0.0}{0.0}{2.7} & \trip{1.0}{40.6}{52.1} & \trip{9.4}{44.8}{52.1} & \trip{0.0}{0.0}{0.0} & \trip{0.0}{0.0}{1.2} & \trip{0.0}{1.0}{3.9} & \trip{0.0}{0.0}{0.0} \\
      WebLINX & \trip{0.0}{0.5}{0.5} & \trip{0.0}{0.5}{0.5} & \trip{0.3}{1.1}{1.6} & \trip{0.0}{0.0}{0.0} & \trip{2.2}{10.9}{13.0} & \trip{2.2}{9.7}{12.9} & \trip{1.3}{35.5}{48.4} & \trip{17.4}{49.0}{57.2} & \trip{0.0}{0.0}{0.0} & \trip{0.0}{0.0}{0.0} & \trip{0.0}{0.2}{1.0} & \trip{0.0}{0.0}{0.0} \\
      GUIAct & \trip{0.0}{0.0}{0.0} & \trip{0.0}{0.0}{0.0} & \trip{0.0}{0.1}{0.1} & \trip{0.0}{0.0}{0.0} & \trip{0.0}{0.5}{1.5} & \trip{0.0}{0.5}{3.5} & \trip{0.5}{17.9}{24.6} & \trip{23.4}{34.9}{39.2} & \trip{0.0}{0.0}{0.0} & \trip{0.0}{0.0}{0.0} & \trip{0.0}{0.1}{0.1} & \trip{0.2}{0.2}{0.4} \\
      \midrule

\rowcolor{blue!30}\multicolumn{13}{c}{\bfseries ShowUI-2B} \\
      \midrule
      Mind2Web & \trip{0.0}{0.0}{0.0} & \trip{0.0}{0.0}{0.0} & \trip{0.0}{0.4}{0.6} & \trip{0.0}{0.6}{1.2} & \trip{1.8}{2.7}{5.3} & \trip{1.0}{3.0}{5.5} & \trip{4.2}{52.9}{69.9} & \trip{4.9}{51.3}{69.3} & \trip{0.0}{0.5}{0.5} & \trip{0.0}{0.0}{0.0} & \trip{0.1}{0.1}{0.5} & \trip{0.0}{0.0}{0.0} \\
      AutoWebGLM & \trip{0.0}{0.0}{0.0} & \trip{0.0}{0.0}{0.0} & \trip{2.0}{4.0}{4.0} & \trip{0.0}{0.0}{2.0} & \trip{0.0}{17.9}{17.9} & \trip{0.0}{12.5}{23.2} & \trip{0.0}{12.0}{16.0} & \trip{0.0}{16.0}{22.0} & \trip{0.0}{0.0}{0.0} & \trip{0.0}{0.0}{0.0} & \trip{0.0}{0.0}{0.0} & \trip{0.0}{0.0}{3.3} \\
      WebArena & \trip{0.0}{0.0}{0.0} & \trip{0.0}{0.0}{0.0} & \trip{0.0}{1.2}{3.6} & \trip{1.2}{2.4}{6.0} & \trip{5.4}{10.8}{27.0} & \trip{2.7}{13.3}{22.7} & \trip{4.2}{40.6}{51.0} & \trip{13.5}{43.8}{51.0} & \trip{0.0}{0.0}{0.0} & \trip{0.0}{0.0}{0.0} & \trip{0.0}{1.9}{5.8} & \trip{0.0}{0.0}{0.0} \\
      WebLINX & \trip{0.0}{0.0}{0.5} & \trip{0.0}{0.0}{0.5} & \trip{0.0}{0.3}{0.5} & \trip{0.3}{0.5}{1.3} & \trip{2.2}{8.7}{10.9} & \trip{2.2}{8.6}{11.8} & \trip{3.5}{36.9}{46.8} & \trip{17.5}{46.0}{56.2} & \trip{0.0}{0.0}{0.0} & \trip{0.0}{0.0}{0.0} & \trip{0.0}{0.2}{0.9} & \trip{0.0}{0.0}{0.0} \\
      GUIAct & \trip{0.0}{0.0}{0.0} & \trip{0.0}{0.0}{0.0} & \trip{0.0}{0.1}{0.1} & \trip{0.0}{0.0}{0.1} & \trip{0.0}{0.0}{0.5} & \trip{0.0}{0.0}{0.0} & \trip{0.9}{18.4}{23.4} & \trip{22.9}{35.1}{38.4} & \trip{0.0}{0.0}{0.0} & \trip{0.0}{0.0}{0.0} & \trip{0.0}{0.1}{0.1} & \trip{0.0}{0.0}{0.0} \\
      \midrule

\rowcolor{blue!30}\multicolumn{13}{c}{\bfseries UI-TARS-2B-SFT} \\
      \midrule
      Mind2Web & \trip{0.0}{0.0}{0.5} & \trip{0.0}{0.0}{0.5} & \trip{0.0}{0.4}{0.6} & \trip{0.0}{0.4}{0.6} & \trip{0.0}{3.5}{6.2} & \trip{2.5}{3.5}{5.5} & \trip{1.5}{51.6}{66.6} & \trip{3.7}{50.9}{66.5} & \trip{0.0}{0.0}{0.0} & \trip{0.0}{0.0}{0.0} & \trip{0.0}{0.1}{0.2} & \trip{0.0}{0.0}{0.7} \\
      AutoWebGLM & \trip{0.0}{0.0}{0.0} & \trip{0.0}{0.0}{0.0} & \trip{0.0}{4.0}{4.0} & \trip{0.0}{4.0}{4.0} & \trip{0.0}{7.1}{14.3} & \trip{0.0}{7.1}{10.7} & \trip{0.0}{14.0}{18.0} & \trip{0.0}{14.0}{18.0} & \trip{0.0}{0.0}{0.0} & \trip{0.0}{2.0}{2.0} & \trip{0.0}{1.8}{3.6} & \trip{0.0}{0.0}{0.0} \\
      WebArena & \trip{0.0}{1.4}{1.4} & \trip{0.0}{0.0}{0.0} & \trip{0.0}{1.2}{2.4} & \trip{0.0}{0.0}{0.0} & \trip{2.7}{8.1}{13.5} & \trip{2.7}{12.0}{22.7} & \trip{1.0}{36.5}{46.9} & \trip{8.3}{39.6}{51.0} & \trip{0.0}{0.0}{0.0} & \trip{0.0}{0.0}{0.0} & \trip{1.0}{1.0}{3.9} & \trip{0.0}{0.0}{0.0} \\
      WebLINX & \trip{0.0}{0.0}{1.0} & \trip{0.0}{0.0}{0.0} & \trip{0.3}{0.8}{1.1} & \trip{0.0}{0.5}{1.1} & \trip{4.3}{8.7}{10.9} & \trip{4.3}{10.8}{17.2} & \trip{2.4}{36.0}{47.3} & \trip{16.6}{47.0}{57.2} & \trip{0.0}{0.0}{0.0} & \trip{0.0}{0.0}{0.0} & \trip{0.0}{0.5}{0.7} & \trip{0.0}{0.0}{0.0} \\
      GUIAct & \trip{0.0}{0.0}{0.0} & \trip{0.0}{0.0}{0.0} & \trip{0.0}{0.0}{0.0} & \trip{0.0}{0.0}{0.0} & \trip{0.0}{0.0}{0.5} & \trip{0.0}{0.0}{0.5} & \trip{0.1}{19.4}{24.9} & \trip{23.8}{36.0}{39.4} & \trip{0.0}{0.0}{0.0} & \trip{0.0}{0.0}{0.0} & \trip{0.0}{0.0}{0.1} & \trip{0.0}{0.0}{0.0} \\
      \midrule

\rowcolor{blue!30}\multicolumn{13}{c}{\bfseries TongUI-3B} \\
      \midrule
      Mind2Web & \trip{0.0}{0.0}{0.0} & \trip{0.0}{0.0}{0.0} & \trip{0.0}{0.1}{0.1} & \trip{0.0}{0.3}{0.3} & \trip{0.0}{1.8}{3.5} & \trip{0.0}{0.0}{1.0} & \trip{5.4}{38.1}{48.9} & \trip{5.5}{39.2}{50.0} & \trip{0.0}{0.0}{0.0} & \trip{0.0}{0.5}{0.5} & \trip{0.0}{0.1}{0.3} & \trip{0.0}{0.0}{0.7} \\
      AutoWebGLM & \trip{0.0}{2.0}{2.0} & \trip{0.0}{0.0}{0.0} & \trip{2.0}{2.0}{4.0} & \trip{0.0}{2.0}{4.0} & \trip{3.6}{14.3}{28.6} & \trip{3.6}{12.5}{21.4} & \trip{0.0}{10.0}{14.0} & \trip{0.0}{8.0}{18.0} & \trip{0.0}{0.0}{0.0} & \trip{2.0}{2.0}{2.0} & \trip{0.0}{0.0}{1.8} & \trip{0.0}{3.3}{3.3} \\
      WebArena & \trip{0.0}{0.0}{0.0} & \trip{0.0}{0.0}{1.4} & \trip{0.0}{1.2}{3.6} & \trip{0.0}{0.0}{1.2} & \trip{0.0}{5.4}{13.5} & \trip{0.0}{2.7}{8.0} & \trip{7.3}{35.4}{49.0} & \trip{6.2}{32.3}{41.7} & \trip{0.0}{0.0}{1.2} & \trip{0.0}{0.0}{0.0} & \trip{0.0}{0.0}{1.9} & \trip{0.0}{0.0}{0.0} \\
      WebLINX & \trip{0.0}{0.0}{0.0} & \trip{0.0}{0.0}{0.5} & \trip{0.0}{0.0}{0.3} & \trip{0.0}{0.0}{0.0} & \trip{4.3}{4.3}{8.7} & \trip{3.2}{4.3}{7.5} & \trip{4.9}{35.4}{46.0} & \trip{16.1}{47.5}{57.2} & \trip{0.0}{0.0}{0.0} & \trip{0.0}{0.0}{0.0} & \trip{0.2}{0.9}{0.9} & \trip{0.0}{0.0}{0.0} \\
      GUIAct & \trip{0.0}{0.0}{0.0} & \trip{0.0}{0.0}{0.0} & \trip{0.0}{0.0}{0.1} & \trip{0.0}{0.1}{0.1} & \trip{0.0}{0.5}{0.5} & \trip{0.0}{0.0}{0.0} & \trip{2.7}{18.8}{24.1} & \trip{20.0}{34.9}{39.0} & \trip{0.0}{0.0}{0.0} & \trip{0.0}{0.0}{0.0} & \trip{0.0}{0.1}{0.1} & \trip{0.0}{0.2}{0.2} \\
      \midrule

\rowcolor{blue!30}\multicolumn{13}{c}{\bfseries GAE-Retriever} \\
      \midrule
      Mind2Web & \trip{0.0}{1.0}{3.5} & \trip{0.0}{0.0}{0.5} & \trip{14.8}{60.0}{84.3} & \trip{4.7}{38.7}{71.1} & \trip{75.2}{95.6}{99.1} & \trip{46.5}{80.5}{92.0} & \trip{9.9}{56.1}{78.5} & \trip{9.8}{49.6}{74.0} & \trip{2.0}{8.5}{13.0} & \trip{1.0}{7.0}{13.0} & \trip{16.2}{45.8}{66.5} & \trip{17.2}{54.5}{70.9} \\
      AutoWebGLM & \trip{12.0}{48.0}{56.0} & \trip{2.0}{16.0}{32.0} & \trip{12.0}{64.0}{82.0} & \trip{2.0}{38.0}{62.0} & \trip{85.7}{96.4}{100.0} & \trip{55.4}{82.1}{91.1} & \trip{4.0}{46.0}{70.0} & \trip{6.0}{42.0}{58.0} & \trip{4.0}{22.0}{40.0} & \trip{0.0}{6.0}{14.0} & \trip{21.4}{78.6}{91.1} & \trip{20.0}{56.7}{70.0} \\
      WebArena & \trip{1.4}{1.4}{9.6} & \trip{0.0}{1.4}{4.1} & \trip{9.5}{21.4}{40.5} & \trip{7.1}{21.4}{36.9} & \trip{51.4}{83.8}{94.6} & \trip{28.0}{72.0}{88.0} & \trip{6.2}{36.5}{51.0} & \trip{5.2}{41.7}{55.2} & \trip{2.4}{10.7}{13.1} & \trip{0.0}{2.4}{9.5} & \trip{13.6}{45.6}{65.0} & \trip{13.3}{31.1}{44.4} \\
      WebLINX & \trip{0.0}{1.5}{5.0} & \trip{0.0}{0.5}{1.0} & \trip{13.6}{56.5}{70.1} & \trip{3.5}{36.8}{59.7} & \trip{80.4}{91.3}{95.7} & \trip{59.1}{84.9}{89.2} & \trip{7.5}{43.6}{56.8} & \trip{13.9}{43.9}{62.7} & \trip{4.0}{8.0}{12.0} & \trip{1.0}{2.5}{11.5} & \trip{22.5}{58.3}{72.3} & \trip{0.0}{4.0}{10.0} \\
      GUIAct & \trip{30.5}{66.5}{82.5} & \trip{2.5}{13.0}{27.5} & \trip{16.4}{42.0}{50.4} & \trip{17.9}{43.0}{49.6} & \trip{95.0}{99.0}{99.5} & \trip{55.0}{85.0}{95.5} & \trip{10.5}{39.2}{47.0} & \trip{17.0}{45.5}{51.6} & \trip{15.5}{58.5}{74.0} & \trip{4.0}{24.5}{40.5} & \trip{29.4}{60.2}{70.5} & \trip{4.7}{13.0}{17.4} \\
      \bottomrule
    \end{tabular}
  \end{adjustbox}
  \label{tab:subtask_recall_ind}
\end{table}


\begin{table}[ht]
  \centering
  \small
  \setlength{\abovecaptionskip}{5pt}
  \setlength{\belowcaptionskip}{0pt}
  \caption{Recall@1/5/10 for all methods on each subtask in the out-of-domain setting.}
  \begin{adjustbox}{max width=\columnwidth}
    \begin{tabular}{@{}l *{12}{c}@{}}
      \toprule
      \textbf{Source}
        & \textbf{$(q,\tau_{1:i})\to\tau_{i+1:n}$}
        & \textbf{$(q,\tau_{i+1:n})\to\tau_{1:i}$}
        & \textbf{$(q,\tau_{1:i})\to s_{i+1}$}
        & \textbf{$(q,\tau_{i+1:n})\to s_i$}
        & \textbf{$q\to\tau_{\equiv}$}
        & \textbf{$q\to\tau_{\sim}$}
        & \textbf{$(q,s_i)\to s_{i+1}$}
        & \textbf{$(q,s_{i+1})\to s_i$}
        & \textbf{$(q,s_i)\to\tau_{i+1:n}$}
        & \textbf{$(q,s_{i+1})\to\tau_{1:i}$}
        & \textbf{$q\to s_i$}
        & \textbf{$q\to s_n$} \\
      \midrule

      \rowcolor{blue!30}\multicolumn{13}{c}{\bfseries Qwen2-VL-2B} \\
      \midrule
      Mind2Web & \trip{0.0}{0.3}{1.0} & \trip{0.2}{1.6}{2.8} & \trip{0.2}{0.7}{1.2} & \trip{0.2}{0.2}{0.3} & \trip{0.8}{4.0}{6.3} & \trip{0.8}{4.6}{7.8} & \trip{2.3}{62.5}{70.7} & \trip{2.4}{61.8}{71.2} & \trip{0.2}{0.3}{0.7} & \trip{0.0}{0.0}{0.3} & \trip{0.1}{0.7}{1.6} & \trip{0.0}{0.0}{0.0} \\
      AutoWebGLM & \trip{0.0}{0.0}{0.0} & \trip{0.0}{0.0}{0.0} & \trip{0.0}{7.5}{7.5} & \trip{0.0}{2.5}{12.5} & \trip{0.0}{5.0}{5.0} & \trip{0.0}{0.0}{4.0} & \trip{0.0}{22.5}{30.0} & \trip{0.0}{27.5}{35.0} & \trip{2.5}{5.0}{5.0} & \trip{0.0}{2.5}{2.5} & \trip{1.7}{3.3}{5.0} & \trip{5.0}{10.0}{10.0} \\
      WebArena & \trip{0.0}{1.7}{3.4} & \trip{0.0}{1.7}{5.2} & \trip{0.0}{0.0}{0.0} & \trip{0.0}{0.0}{0.0} & \trip{5.0}{5.0}{15.0} & \trip{5.0}{6.0}{15.0} & \trip{1.7}{20.7}{24.1} & \trip{3.4}{34.5}{39.7} & \trip{0.0}{1.7}{1.7} & \trip{0.0}{0.0}{0.0} & \trip{0.0}{0.0}{0.0} & \trip{0.0}{0.0}{0.0} \\
      WebLINX & \trip{0.0}{0.0}{0.9} & \trip{0.9}{0.9}{0.9} & \trip{0.9}{0.9}{0.9} & \trip{0.0}{0.0}{0.0} & \trip{5.0}{10.0}{25.0} & \trip{6.0}{14.0}{24.0} & \trip{0.9}{47.8}{50.4} & \trip{10.4}{49.6}{53.0} & \trip{0.0}{0.0}{0.0} & \trip{0.0}{0.0}{0.0} & \trip{0.0}{0.0}{0.0} & \trip{0.0}{0.0}{0.0} \\
      GUIAct & \trip{0.0}{0.0}{0.0} & \trip{0.0}{0.0}{0.1} & \trip{0.0}{0.1}{0.1} & \trip{0.0}{0.1}{0.2} & \trip{0.4}{3.5}{5.6} & \trip{0.5}{2.4}{4.2} & \trip{0.5}{27.6}{34.5} & \trip{23.7}{40.4}{44.6} & \trip{0.1}{0.1}{0.1} & \trip{0.0}{0.0}{0.0} & \trip{0.0}{0.0}{0.1} & \trip{0.0}{0.0}{0.0} \\
      \midrule

\rowcolor{blue!30}\multicolumn{13}{c}{\bfseries Qwen2.5-VL-3B} \\
      \midrule
      Mind2Web & \trip{0.0}{0.3}{0.3} & \trip{0.0}{0.0}{0.0} & \trip{0.0}{0.2}{0.2} & \trip{0.0}{0.2}{0.2} & \trip{0.0}{0.0}{0.0} & \trip{0.0}{0.5}{1.4} & \trip{2.3}{31.8}{37.2} & \trip{5.9}{34.2}{38.7} & \trip{0.0}{0.2}{0.2} & \trip{0.0}{0.0}{0.2} & \trip{0.0}{0.0}{0.1} & \trip{0.0}{0.0}{0.0} \\
      AutoWebGLM & \trip{0.0}{0.0}{0.0} & \trip{0.0}{0.0}{0.0} & \trip{2.5}{2.5}{2.5} & \trip{0.0}{2.5}{2.5} & \trip{5.0}{15.0}{35.0} & \trip{4.0}{12.0}{21.0} & \trip{0.0}{5.0}{5.0} & \trip{0.0}{5.0}{5.0} & \trip{0.0}{0.0}{2.5} & \trip{0.0}{0.0}{0.0} & \trip{1.7}{1.7}{5.0} & \trip{0.0}{0.0}{0.0} \\
      WebArena & \trip{0.0}{1.7}{1.7} & \trip{0.0}{0.0}{0.0} & \trip{0.0}{0.0}{6.9} & \trip{0.0}{0.0}{6.9} & \trip{0.0}{5.0}{15.0} & \trip{0.0}{6.0}{10.0} & \trip{3.4}{12.1}{13.8} & \trip{1.7}{13.8}{20.7} & \trip{0.0}{0.0}{0.0} & \trip{0.0}{0.0}{0.0} & \trip{0.0}{0.0}{4.1} & \trip{0.0}{0.0}{0.0} \\
      WebLINX & \trip{0.0}{0.0}{0.9} & \trip{0.0}{0.0}{0.0} & \trip{0.0}{0.0}{0.0} & \trip{0.0}{0.0}{0.0} & \trip{0.0}{0.0}{5.0} & \trip{0.0}{3.0}{8.0} & \trip{5.2}{38.3}{43.5} & \trip{12.2}{43.5}{47.8} & \trip{0.0}{0.0}{0.0} & \trip{0.0}{0.0}{0.0} & \trip{0.0}{0.8}{0.8} & \trip{0.0}{0.0}{5.0} \\
      GUIAct & \trip{0.0}{0.0}{0.0} & \trip{0.0}{0.0}{0.0} & \trip{0.0}{0.0}{0.2} & \trip{0.1}{0.1}{0.2} & \trip{0.0}{1.4}{4.2} & \trip{0.7}{1.6}{3.3} & \trip{2.3}{29.0}{36.5} & \trip{21.1}{41.0}{47.7} & \trip{0.1}{0.5}{0.7} & \trip{0.0}{0.0}{0.0} & \trip{0.1}{0.2}{0.2} & \trip{0.0}{0.0}{0.0} \\
      \midrule

\rowcolor{blue!30}\multicolumn{13}{c}{\bfseries LamRA-Ret} \\
      \midrule
      Mind2Web & \trip{0.0}{0.3}{0.9} & \trip{0.0}{0.3}{1.0} & \trip{0.2}{1.0}{2.1} & \trip{0.3}{0.7}{1.7} & \trip{4.8}{14.3}{20.6} & \trip{2.1}{9.0}{15.4} & \trip{2.1}{54.5}{62.8} & \trip{3.1}{55.8}{63.5} & \trip{0.0}{0.3}{0.7} & \trip{0.0}{0.0}{0.0} & \trip{1.2}{5.8}{8.6} & \trip{2.4}{5.6}{7.1} \\
      AutoWebGLM & \trip{0.0}{0.0}{0.0} & \trip{0.0}{0.0}{0.0} & \trip{0.0}{5.0}{5.0} & \trip{5.0}{10.0}{15.0} & \trip{30.0}{65.0}{70.0} & \trip{21.0}{43.0}{60.0} & \trip{0.0}{25.0}{27.5} & \trip{0.0}{25.0}{32.5} & \trip{10.0}{17.5}{22.5} & \trip{0.0}{2.5}{2.5} & \trip{1.7}{8.3}{23.3} & \trip{0.0}{10.0}{20.0} \\
      WebArena & \trip{0.0}{0.0}{3.4} & \trip{0.0}{3.4}{5.2} & \trip{0.0}{0.0}{0.0} & \trip{0.0}{0.0}{0.0} & \trip{10.0}{35.0}{35.0} & \trip{6.0}{18.0}{27.0} & \trip{0.0}{19.0}{27.6} & \trip{5.2}{27.6}{32.8} & \trip{0.0}{0.0}{0.0} & \trip{0.0}{0.0}{0.0} & \trip{2.7}{6.8}{8.2} & \trip{5.0}{5.0}{5.0} \\
      WebLINX & \trip{0.0}{0.0}{0.0} & \trip{0.0}{0.0}{0.0} & \trip{0.0}{2.6}{4.3} & \trip{0.0}{2.6}{6.1} & \trip{15.0}{50.0}{55.0} & \trip{4.0}{30.0}{48.0} & \trip{1.7}{48.7}{53.9} & \trip{13.0}{51.3}{56.5} & \trip{0.0}{0.0}{0.0} & \trip{0.9}{0.9}{0.9} & \trip{5.1}{15.3}{19.5} & \trip{0.0}{10.0}{15.0} \\
      GUIAct & \trip{0.0}{0.0}{0.0} & \trip{0.0}{0.0}{0.0} & \trip{0.1}{0.2}{0.2} & \trip{0.3}{0.7}{1.2} & \trip{27.8}{51.8}{60.6} & \trip{12.2}{30.4}{38.5} & \trip{0.7}{26.6}{33.4} & \trip{23.3}{40.3}{44.7} & \trip{0.2}{0.7}{1.8} & \trip{0.0}{0.0}{0.0} & \trip{0.6}{1.8}{3.2} & \trip{0.4}{1.8}{1.8} \\
      \midrule

\rowcolor{blue!30}\multicolumn{13}{c}{\bfseries ColQwen2-v1.0} \\
      \midrule
      Mind2Web & \trip{0.0}{1.6}{3.1} & \trip{0.0}{1.6}{4.0} & \trip{4.0}{13.4}{20.2} & \trip{1.0}{5.8}{11.0} & \trip{20.6}{38.9}{49.2} & \trip{13.0}{26.0}{36.8} & \trip{1.2}{63.2}{77.0} & \trip{2.4}{66.3}{77.3} & \trip{1.9}{6.6}{11.2} & \trip{2.8}{11.0}{18.8} & \trip{5.4}{18.0}{25.3} & \trip{11.1}{21.4}{27.0} \\
      AutoWebGLM & \trip{0.0}{0.0}{0.0} & \trip{0.0}{0.0}{0.0} & \trip{2.5}{5.0}{10.0} & \trip{5.0}{35.0}{45.0} & \trip{10.0}{35.0}{55.0} & \trip{0.0}{20.0}{35.0} & \trip{0.0}{55.0}{62.5} & \trip{0.0}{55.0}{60.0} & \trip{0.0}{7.5}{10.0} & \trip{0.0}{0.0}{0.0} & \trip{13.3}{25.0}{33.3} & \trip{5.0}{15.0}{35.0} \\
      WebArena & \trip{0.0}{0.0}{3.4} & \trip{0.0}{3.4}{3.4} & \trip{0.0}{0.0}{1.7} & \trip{0.0}{1.7}{3.4} & \trip{20.0}{40.0}{55.0} & \trip{11.0}{29.0}{45.0} & \trip{1.7}{20.7}{25.9} & \trip{3.4}{31.0}{43.1} & \trip{0.0}{1.7}{5.2} & \trip{3.4}{6.9}{10.3} & \trip{2.7}{8.2}{12.3} & \trip{0.0}{5.0}{15.0} \\
      WebLINX & \trip{0.0}{0.0}{0.9} & \trip{0.0}{0.9}{0.9} & \trip{3.5}{12.2}{17.4} & \trip{1.7}{6.1}{6.1} & \trip{20.0}{50.0}{70.0} & \trip{16.0}{41.0}{51.0} & \trip{2.6}{48.7}{52.2} & \trip{10.4}{54.8}{58.3} & \trip{0.0}{0.9}{0.9} & \trip{0.9}{7.8}{9.6} & \trip{4.2}{11.0}{12.7} & \trip{5.0}{15.0}{15.0} \\
      GUIAct & \trip{0.1}{0.3}{1.6} & \trip{0.0}{0.4}{2.2} & \trip{1.0}{3.1}{4.1} & \trip{2.7}{8.2}{11.2} & \trip{71.1}{85.6}{90.1} & \trip{28.9}{56.1}{67.5} & \trip{0.5}{34.1}{42.0} & \trip{23.5}{46.6}{53.2} & \trip{1.6}{6.5}{12.2} & \trip{0.6}{2.4}{3.8} & \trip{0.6}{2.9}{4.7} & \trip{2.1}{7.0}{10.2} \\
      \midrule

\rowcolor{blue!30}\multicolumn{13}{c}{\bfseries GME-Qwen2VL-2B} \\
      \midrule
      Mind2Web & \trip{0.0}{2.1}{6.6} & \trip{0.0}{1.7}{6.5} & \trip{4.2}{18.5}{25.3} & \trip{4.2}{14.8}{21.3} & \trip{30.2}{57.1}{73.0} & \trip{16.5}{37.1}{47.6} & \trip{2.4}{65.3}{75.4} & \trip{2.3}{66.5}{76.8} & \trip{3.0}{13.4}{27.9} & \trip{3.3}{13.1}{26.5} & \trip{5.5}{15.4}{21.1} & \trip{4.0}{14.3}{18.3} \\
      AutoWebGLM & \trip{0.0}{0.0}{0.0} & \trip{0.0}{0.0}{0.0} & \trip{5.0}{37.5}{50.0} & \trip{0.0}{25.0}{40.0} & \trip{25.0}{45.0}{60.0} & \trip{19.0}{30.0}{49.0} & \trip{0.0}{70.0}{70.0} & \trip{0.0}{70.0}{72.5} & \trip{0.0}{5.0}{7.5} & \trip{7.5}{40.0}{57.5} & \trip{20.0}{36.7}{48.3} & \trip{20.0}{30.0}{40.0} \\
      WebArena & \trip{0.0}{3.4}{5.2} & \trip{0.0}{6.9}{8.6} & \trip{0.0}{3.4}{8.6} & \trip{0.0}{6.9}{10.3} & \trip{25.0}{70.0}{75.0} & \trip{15.0}{44.0}{65.0} & \trip{5.2}{24.1}{29.3} & \trip{5.2}{29.3}{41.4} & \trip{3.4}{6.9}{20.7} & \trip{1.7}{5.2}{8.6} & \trip{0.0}{4.1}{5.5} & \trip{0.0}{10.0}{20.0} \\
      WebLINX & \trip{0.0}{1.7}{1.7} & \trip{0.0}{2.6}{5.2} & \trip{7.8}{22.6}{34.8} & \trip{5.2}{16.5}{20.0} & \trip{60.0}{85.0}{90.0} & \trip{29.0}{59.0}{71.0} & \trip{2.6}{52.2}{55.7} & \trip{11.3}{53.9}{59.1} & \trip{0.0}{6.1}{9.6} & \trip{0.9}{8.7}{19.1} & \trip{4.2}{7.6}{12.7} & \trip{10.0}{20.0}{20.0} \\
      GUIAct & \trip{0.0}{0.0}{0.0} & \trip{0.0}{0.0}{0.0} & \trip{2.2}{8.6}{11.9} & \trip{3.6}{11.8}{16.1} & \trip{66.2}{87.3}{93.0} & \trip{22.5}{52.2}{64.9} & \trip{1.1}{35.6}{44.2} & \trip{23.3}{47.9}{53.6} & \trip{1.0}{2.6}{4.2} & \trip{2.6}{9.7}{15.4} & \trip{1.2}{3.5}{4.7} & \trip{0.4}{6.3}{10.6} \\
      \midrule

\rowcolor{blue!30}\multicolumn{13}{c}{\bfseries UniSE-MLLM} \\
      \midrule
      Mind2Web & \trip{0.0}{1.9}{4.7} & \trip{0.3}{2.1}{5.1} & \trip{3.3}{8.7}{12.2} & \trip{1.9}{6.3}{10.1} & \trip{4.8}{8.7}{16.7} & \trip{4.1}{9.4}{12.1} & \trip{6.3}{65.4}{74.5} & \trip{2.4}{63.0}{74.3} & \trip{3.0}{10.3}{15.9} & \trip{0.9}{4.2}{8.2} & \trip{4.7}{12.2}{21.4} & \trip{0.8}{3.2}{7.9} \\
      AutoWebGLM & \trip{0.0}{0.0}{0.0} & \trip{0.0}{0.0}{0.0} & \trip{2.5}{10.0}{15.0} & \trip{2.5}{20.0}{22.5} & \trip{5.0}{5.0}{25.0} & \trip{5.0}{8.0}{25.0} & \trip{0.0}{35.0}{42.5} & \trip{0.0}{32.5}{42.5} & \trip{5.0}{15.0}{15.0} & \trip{0.0}{2.5}{2.5} & \trip{0.0}{6.7}{11.7} & \trip{5.0}{5.0}{10.0} \\
      WebArena & \trip{1.7}{3.4}{6.9} & \trip{0.0}{6.9}{6.9} & \trip{0.0}{0.0}{1.7} & \trip{1.7}{1.7}{3.4} & \trip{0.0}{15.0}{25.0} & \trip{0.0}{8.0}{21.0} & \trip{8.6}{31.0}{34.5} & \trip{3.4}{36.2}{41.4} & \trip{0.0}{5.2}{8.6} & \trip{0.0}{0.0}{1.7} & \trip{2.7}{2.7}{2.7} & \trip{0.0}{0.0}{0.0} \\
      WebLINX & \trip{0.0}{0.0}{0.9} & \trip{0.0}{0.9}{4.3} & \trip{0.9}{1.7}{2.6} & \trip{0.9}{0.9}{1.7} & \trip{15.0}{20.0}{65.0} & \trip{7.0}{23.0}{36.0} & \trip{3.5}{49.6}{51.3} & \trip{11.3}{51.3}{55.7} & \trip{1.7}{5.2}{7.0} & \trip{0.0}{0.9}{1.7} & \trip{0.0}{0.0}{0.0} & \trip{0.0}{0.0}{0.0} \\
      GUIAct & \trip{0.0}{0.0}{0.1} & \trip{0.0}{0.0}{0.1} & \trip{0.3}{0.8}{1.2} & \trip{0.5}{1.5}{2.5} & \trip{3.9}{10.2}{14.1} & \trip{1.8}{5.8}{8.6} & \trip{1.6}{31.6}{38.3} & \trip{23.5}{42.6}{48.0} & \trip{0.1}{0.8}{1.1} & \trip{0.0}{0.0}{0.0} & \trip{0.1}{0.3}{0.4} & \trip{0.0}{0.4}{0.7} \\
      \midrule

\rowcolor{blue!30}\multicolumn{13}{c}{\bfseries VLM2Vec-Qwen2VL-2B} \\
      \midrule
      Mind2Web & \trip{0.0}{2.3}{7.0} & \trip{0.0}{2.1}{9.1} & \trip{9.6}{45.2}{62.5} & \trip{6.1}{46.8}{65.6} & \trip{38.9}{69.0}{79.4} & \trip{21.9}{42.5}{55.1} & \trip{1.2}{73.6}{84.1} & \trip{1.6}{75.6}{87.3} & \trip{3.1}{18.5}{33.2} & \trip{1.7}{11.7}{23.4} & \trip{21.4}{56.6}{69.6} & \trip{23.0}{57.1}{65.1} \\
      AutoWebGLM & \trip{0.0}{0.0}{5.0} & \trip{0.0}{2.5}{7.5} & \trip{7.5}{60.0}{72.5} & \trip{7.5}{72.5}{87.5} & \trip{50.0}{90.0}{95.0} & \trip{31.0}{61.0}{75.0} & \trip{0.0}{87.5}{90.0} & \trip{0.0}{85.0}{92.5} & \trip{7.5}{47.5}{70.0} & \trip{5.0}{30.0}{57.5} & \trip{36.7}{70.0}{81.7} & \trip{20.0}{40.0}{45.0} \\
      WebArena & \trip{0.0}{5.2}{5.2} & \trip{0.0}{5.2}{8.6} & \trip{6.9}{13.8}{17.2} & \trip{1.7}{19.0}{20.7} & \trip{35.0}{70.0}{80.0} & \trip{13.0}{43.0}{65.0} & \trip{3.4}{27.6}{31.0} & \trip{1.7}{34.5}{48.3} & \trip{1.7}{17.2}{27.6} & \trip{0.0}{5.2}{13.8} & \trip{12.3}{27.4}{37.0} & \trip{5.0}{15.0}{15.0} \\
      WebLINX & \trip{0.0}{0.9}{6.1} & \trip{0.0}{1.7}{5.2} & \trip{8.7}{28.7}{42.6} & \trip{7.0}{30.4}{41.7} & \trip{70.0}{90.0}{100.0} & \trip{29.0}{62.0}{83.0} & \trip{4.3}{60.9}{66.1} & \trip{10.4}{58.3}{67.8} & \trip{0.0}{5.2}{16.5} & \trip{0.9}{8.7}{19.1} & \trip{15.3}{36.4}{43.2} & \trip{0.0}{10.0}{15.0} \\
      GUIAct & \trip{0.0}{0.0}{0.3} & \trip{0.0}{0.1}{0.9} & \trip{2.3}{8.7}{12.0} & \trip{5.1}{16.3}{20.7} & \trip{59.9}{79.2}{86.3} & \trip{15.8}{39.7}{54.7} & \trip{0.6}{39.2}{47.7} & \trip{23.6}{50.4}{58.0} & \trip{10.8}{32.8}{45.8} & \trip{0.1}{1.7}{4.9} & \trip{4.8}{12.6}{16.2} & \trip{2.1}{5.6}{6.7} \\
      \midrule

\rowcolor{blue!30}\multicolumn{13}{c}{\bfseries VLM2Vec-V2.2} \\
      \midrule
      Mind2Web & \trip{0.0}{1.7}{10.8} & \trip{0.0}{3.5}{13.4} & \trip{10.1}{57.6}{77.5} & \trip{2.8}{58.6}{79.2} & \trip{79.4}{90.5}{95.2} & \trip{40.3}{67.8}{78.7} & \trip{4.2}{73.6}{85.3} & \trip{2.3}{73.3}{86.4} & \trip{2.3}{16.2}{33.7} & \trip{1.7}{11.7}{29.5} & \trip{27.7}{67.1}{79.6} & \trip{37.3}{76.2}{83.3} \\
      AutoWebGLM & \trip{0.0}{20.0}{60.0} & \trip{0.0}{22.5}{65.0} & \trip{0.0}{85.0}{92.5} & \trip{0.0}{90.0}{95.0} & \trip{75.0}{85.0}{95.0} & \trip{46.0}{67.0}{77.0} & \trip{0.0}{87.5}{97.5} & \trip{0.0}{92.5}{95.0} & \trip{2.5}{47.5}{77.5} & \trip{0.0}{32.5}{67.5} & \trip{56.7}{86.7}{91.7} & \trip{20.0}{60.0}{60.0} \\
      WebArena & \trip{0.0}{8.6}{10.3} & \trip{0.0}{10.3}{12.1} & \trip{1.7}{13.8}{24.1} & \trip{3.4}{29.3}{44.8} & \trip{55.0}{75.0}{85.0} & \trip{31.0}{60.0}{75.0} & \trip{10.3}{37.9}{44.8} & \trip{6.9}{34.5}{44.8} & \trip{3.4}{19.0}{25.9} & \trip{0.0}{5.2}{19.0} & \trip{20.5}{42.5}{54.8} & \trip{25.0}{45.0}{55.0} \\
      WebLINX & \trip{0.0}{0.9}{6.1} & \trip{0.0}{0.0}{3.5} & \trip{7.0}{65.2}{78.3} & \trip{3.5}{50.4}{67.0} & \trip{90.0}{100.0}{100.0} & \trip{55.0}{87.0}{95.0} & \trip{6.1}{67.8}{73.9} & \trip{12.2}{65.2}{72.2} & \trip{4.3}{10.4}{20.0} & \trip{2.6}{9.6}{18.3} & \trip{36.4}{77.1}{89.0} & \trip{30.0}{65.0}{70.0} \\
      GUIAct & \trip{0.0}{0.1}{1.4} & \trip{0.0}{0.1}{1.9} & \trip{8.7}{35.6}{44.7} & \trip{9.9}{39.3}{49.1} & \trip{96.5}{99.3}{99.6} & \trip{39.7}{72.4}{83.7} & \trip{1.8}{42.5}{50.0} & \trip{22.3}{52.8}{60.7} & \trip{6.4}{24.0}{34.1} & \trip{0.6}{5.7}{10.9} & \trip{28.2}{58.3}{68.4} & \trip{10.6}{38.0}{41.9} \\
      \midrule

\rowcolor{blue!30}\multicolumn{13}{c}{\bfseries UGround-V1-2B} \\
      \midrule
      Mind2Web & \trip{0.0}{0.2}{0.5} & \trip{0.2}{0.7}{1.0} & \trip{0.3}{0.5}{1.2} & \trip{0.2}{0.9}{1.0} & \trip{0.8}{4.8}{7.9} & \trip{0.8}{4.3}{8.1} & \trip{2.6}{50.1}{57.6} & \trip{2.1}{50.1}{57.6} & \trip{0.0}{0.2}{0.5} & \trip{0.0}{0.2}{0.5} & \trip{0.3}{0.6}{0.7} & \trip{0.0}{0.0}{0.0} \\
      AutoWebGLM & \trip{0.0}{0.0}{0.0} & \trip{0.0}{0.0}{0.0} & \trip{0.0}{0.0}{2.5} & \trip{0.0}{0.0}{0.0} & \trip{0.0}{0.0}{25.0} & \trip{0.0}{9.0}{30.0} & \trip{0.0}{12.5}{20.0} & \trip{0.0}{15.0}{20.0} & \trip{0.0}{0.0}{2.5} & \trip{0.0}{0.0}{0.0} & \trip{0.0}{1.7}{5.0} & \trip{0.0}{5.0}{5.0} \\
      WebArena & \trip{0.0}{0.0}{0.0} & \trip{0.0}{0.0}{1.7} & \trip{3.4}{3.4}{3.4} & \trip{0.0}{1.7}{1.7} & \trip{5.0}{25.0}{45.0} & \trip{5.0}{23.0}{43.0} & \trip{0.0}{19.0}{25.9} & \trip{1.7}{27.6}{34.5} & \trip{0.0}{0.0}{0.0} & \trip{0.0}{0.0}{0.0} & \trip{0.0}{1.4}{1.4} & \trip{0.0}{0.0}{0.0} \\
      WebLINX & \trip{0.0}{0.0}{0.0} & \trip{0.0}{0.0}{0.0} & \trip{0.0}{0.0}{0.0} & \trip{0.0}{0.0}{0.0} & \trip{0.0}{0.0}{15.0} & \trip{0.0}{6.0}{15.0} & \trip{2.6}{47.8}{49.6} & \trip{12.2}{50.4}{53.9} & \trip{0.0}{0.0}{0.0} & \trip{0.9}{0.9}{0.9} & \trip{0.0}{0.8}{0.8} & \trip{0.0}{0.0}{0.0} \\
      GUIAct & \trip{0.0}{0.0}{0.0} & \trip{0.0}{0.0}{0.0} & \trip{0.1}{0.4}{0.5} & \trip{0.1}{0.1}{0.4} & \trip{0.0}{2.1}{3.9} & \trip{0.4}{1.8}{3.8} & \trip{0.6}{27.8}{35.9} & \trip{23.4}{40.4}{46.8} & \trip{0.0}{0.0}{0.0} & \trip{0.1}{0.2}{0.4} & \trip{0.1}{0.1}{0.1} & \trip{0.0}{0.0}{0.0} \\
      \midrule

\rowcolor{blue!30}\multicolumn{13}{c}{\bfseries ShowUI-2B} \\
      \midrule
      Mind2Web & \trip{0.0}{0.5}{0.9} & \trip{0.0}{0.7}{0.9} & \trip{0.0}{0.0}{0.0} & \trip{0.0}{0.2}{0.2} & \trip{0.0}{1.6}{4.8} & \trip{0.0}{1.6}{3.8} & \trip{3.3}{58.6}{67.4} & \trip{3.1}{55.8}{65.4} & \trip{0.0}{0.2}{0.2} & \trip{0.0}{0.0}{0.0} & \trip{0.3}{0.7}{0.9} & \trip{0.0}{0.8}{0.8} \\
      AutoWebGLM & \trip{0.0}{0.0}{0.0} & \trip{0.0}{0.0}{0.0} & \trip{0.0}{2.5}{2.5} & \trip{0.0}{2.5}{7.5} & \trip{5.0}{15.0}{25.0} & \trip{5.0}{15.0}{20.0} & \trip{0.0}{17.5}{17.5} & \trip{0.0}{17.5}{17.5} & \trip{0.0}{0.0}{0.0} & \trip{0.0}{0.0}{2.5} & \trip{1.7}{3.3}{5.0} & \trip{5.0}{5.0}{5.0} \\
      WebArena & \trip{0.0}{0.0}{0.0} & \trip{0.0}{0.0}{1.7} & \trip{0.0}{1.7}{1.7} & \trip{0.0}{1.7}{1.7} & \trip{0.0}{5.0}{5.0} & \trip{1.0}{2.0}{7.0} & \trip{1.7}{24.1}{29.3} & \trip{1.7}{31.0}{34.5} & \trip{0.0}{0.0}{0.0} & \trip{0.0}{0.0}{0.0} & \trip{0.0}{0.0}{1.4} & \trip{0.0}{0.0}{0.0} \\
      WebLINX & \trip{0.0}{0.0}{0.0} & \trip{0.0}{0.9}{1.7} & \trip{0.0}{0.0}{0.0} & \trip{0.0}{0.9}{2.6} & \trip{0.0}{5.0}{15.0} & \trip{0.0}{2.0}{17.0} & \trip{5.2}{47.8}{49.6} & \trip{11.3}{46.1}{52.2} & \trip{0.0}{0.0}{0.0} & \trip{0.0}{0.0}{0.0} & \trip{0.0}{0.0}{0.0} & \trip{0.0}{0.0}{0.0} \\
      GUIAct & \trip{0.0}{0.0}{0.0} & \trip{0.0}{0.0}{0.0} & \trip{0.0}{0.1}{0.1} & \trip{0.0}{0.1}{0.2} & \trip{0.4}{2.5}{3.5} & \trip{0.4}{2.7}{4.2} & \trip{0.7}{27.2}{33.8} & \trip{23.6}{40.2}{44.6} & \trip{0.0}{0.1}{0.1} & \trip{0.1}{0.1}{0.1} & \trip{0.0}{0.0}{0.0} & \trip{0.0}{0.0}{0.4} \\
      \midrule

\rowcolor{blue!30}\multicolumn{13}{c}{\bfseries UI-TARS-2B-SFT} \\
      \midrule
      Mind2Web & \trip{0.2}{0.5}{0.7} & \trip{0.0}{0.0}{0.2} & \trip{0.0}{0.2}{0.2} & \trip{0.0}{0.0}{0.7} & \trip{0.8}{0.8}{1.6} & \trip{0.5}{0.6}{2.5} & \trip{1.0}{51.3}{57.8} & \trip{3.3}{50.6}{58.3} & \trip{0.0}{0.0}{0.0} & \trip{0.0}{0.2}{0.3} & \trip{0.0}{0.1}{0.1} & \trip{0.0}{0.8}{0.8} \\
      AutoWebGLM & \trip{0.0}{0.0}{0.0} & \trip{0.0}{0.0}{0.0} & \trip{0.0}{2.5}{5.0} & \trip{2.5}{2.5}{5.0} & \trip{5.0}{5.0}{35.0} & \trip{5.0}{13.0}{32.0} & \trip{0.0}{15.0}{20.0} & \trip{0.0}{10.0}{12.5} & \trip{0.0}{0.0}{0.0} & \trip{0.0}{0.0}{0.0} & \trip{0.0}{0.0}{0.0} & \trip{0.0}{0.0}{0.0} \\
      WebArena & \trip{0.0}{0.0}{1.7} & \trip{0.0}{1.7}{3.4} & \trip{0.0}{1.7}{3.4} & \trip{0.0}{0.0}{0.0} & \trip{0.0}{0.0}{25.0} & \trip{0.0}{4.0}{14.0} & \trip{3.4}{19.0}{24.1} & \trip{5.2}{25.9}{34.5} & \trip{0.0}{0.0}{0.0} & \trip{0.0}{0.0}{0.0} & \trip{0.0}{0.0}{0.0} & \trip{0.0}{0.0}{0.0} \\
      WebLINX & \trip{0.0}{0.0}{0.0} & \trip{0.0}{0.0}{0.0} & \trip{0.0}{0.9}{0.9} & \trip{0.0}{0.9}{0.9} & \trip{0.0}{0.0}{5.0} & \trip{0.0}{2.0}{8.0} & \trip{0.9}{44.3}{48.7} & \trip{13.0}{47.0}{54.8} & \trip{0.0}{0.0}{0.0} & \trip{0.0}{0.0}{0.0} & \trip{0.0}{0.0}{0.0} & \trip{0.0}{0.0}{0.0} \\
      GUIAct & \trip{0.0}{0.0}{0.1} & \trip{0.0}{0.0}{0.0} & \trip{0.1}{0.1}{0.1} & \trip{0.1}{0.3}{0.3} & \trip{1.4}{2.8}{3.9} & \trip{0.4}{2.7}{3.7} & \trip{0.1}{27.6}{34.3} & \trip{23.6}{39.8}{44.7} & \trip{0.1}{0.1}{0.2} & \trip{0.0}{0.0}{0.0} & \trip{0.0}{0.0}{0.0} & \trip{0.0}{0.0}{0.0} \\
      \midrule

\rowcolor{blue!30}\multicolumn{13}{c}{\bfseries TongUI-3B} \\
      \midrule
      Mind2Web & \trip{0.0}{0.0}{0.0} & \trip{0.0}{0.2}{0.5} & \trip{0.2}{0.3}{0.7} & \trip{0.0}{0.0}{0.2} & \trip{0.8}{3.2}{5.6} & \trip{1.0}{3.2}{5.7} & \trip{4.0}{35.3}{39.8} & \trip{4.4}{36.8}{39.8} & \trip{0.0}{0.0}{0.0} & \trip{0.0}{0.0}{0.0} & \trip{0.3}{0.9}{0.9} & \trip{0.8}{0.8}{0.8} \\
      AutoWebGLM & \trip{0.0}{0.0}{0.0} & \trip{0.0}{0.0}{0.0} & \trip{0.0}{2.5}{5.0} & \trip{0.0}{0.0}{5.0} & \trip{0.0}{5.0}{15.0} & \trip{0.0}{8.0}{19.0} & \trip{0.0}{5.0}{7.5} & \trip{0.0}{0.0}{0.0} & \trip{0.0}{2.5}{2.5} & \trip{0.0}{2.5}{2.5} & \trip{0.0}{3.3}{5.0} & \trip{0.0}{10.0}{10.0} \\
      WebArena & \trip{0.0}{0.0}{0.0} & \trip{0.0}{0.0}{1.7} & \trip{0.0}{0.0}{0.0} & \trip{0.0}{1.7}{1.7} & \trip{5.0}{15.0}{30.0} & \trip{4.0}{17.0}{28.0} & \trip{5.2}{13.8}{17.2} & \trip{5.2}{19.0}{22.4} & \trip{0.0}{1.7}{1.7} & \trip{0.0}{0.0}{0.0} & \trip{0.0}{2.7}{4.1} & \trip{0.0}{0.0}{0.0} \\
      WebLINX & \trip{0.0}{0.0}{0.9} & \trip{0.0}{0.0}{0.0} & \trip{0.0}{0.0}{0.0} & \trip{0.0}{0.0}{0.0} & \trip{0.0}{15.0}{25.0} & \trip{1.0}{15.0}{31.0} & \trip{7.0}{47.0}{52.2} & \trip{10.4}{46.1}{53.9} & \trip{0.0}{0.0}{0.0} & \trip{0.0}{0.0}{0.0} & \trip{0.0}{0.0}{0.0} & \trip{0.0}{0.0}{0.0} \\
      GUIAct & \trip{0.0}{0.0}{0.0} & \trip{0.0}{0.0}{0.0} & \trip{0.1}{0.1}{0.1} & \trip{0.0}{0.1}{0.1} & \trip{0.4}{2.5}{5.3} & \trip{0.4}{2.5}{4.8} & \trip{3.8}{29.3}{37.1} & \trip{20.2}{41.7}{47.8} & \trip{0.0}{0.0}{0.0} & \trip{0.0}{0.0}{0.0} & \trip{0.0}{0.1}{0.1} & \trip{0.0}{0.0}{0.0} \\
      \midrule

\rowcolor{blue!30}\multicolumn{13}{c}{\bfseries GAE-Retriever} \\
      \midrule
      Mind2Web & \trip{9.4}{31.1}{49.7} & \trip{0.9}{14.8}{30.4} & \trip{21.5}{76.4}{88.5} & \trip{3.3}{59.3}{76.3} & \trip{65.9}{93.7}{97.6} & \trip{32.4}{72.5}{84.3} & \trip{18.8}{77.7}{88.8} & \trip{16.8}{71.7}{88.8} & \trip{9.2}{33.2}{47.5} & \trip{1.4}{24.1}{45.0} & \trip{34.1}{75.0}{85.3} & \trip{46.0}{74.6}{80.2} \\
      AutoWebGLM & \trip{25.0}{77.5}{92.5} & \trip{2.5}{57.5}{82.5} & \trip{12.5}{97.5}{97.5} & \trip{7.5}{97.5}{100.0} & \trip{60.0}{90.0}{95.0} & \trip{45.0}{73.0}{82.0} & \trip{10.0}{87.5}{100.0} & \trip{5.0}{95.0}{95.0} & \trip{40.0}{70.0}{95.0} & \trip{2.5}{57.5}{82.5} & \trip{65.0}{96.7}{100.0} & \trip{45.0}{65.0}{75.0} \\
      WebArena & \trip{10.3}{27.6}{34.5} & \trip{1.7}{13.8}{22.4} & \trip{12.1}{36.2}{50.0} & \trip{5.2}{32.8}{41.4} & \trip{25.0}{50.0}{70.0} & \trip{16.0}{47.0}{74.0} & \trip{13.8}{37.9}{48.3} & \trip{12.1}{41.4}{48.3} & \trip{10.3}{27.6}{39.7} & \trip{0.0}{17.2}{22.4} & \trip{26.0}{58.9}{72.6} & \trip{0.0}{25.0}{40.0} \\
      WebLINX & \trip{7.0}{21.7}{37.4} & \trip{0.0}{4.3}{13.0} & \trip{13.0}{73.9}{77.4} & \trip{5.2}{51.3}{64.3} & \trip{80.0}{100.0}{100.0} & \trip{58.0}{87.0}{98.0} & \trip{17.4}{71.3}{77.4} & \trip{20.9}{66.1}{74.8} & \trip{7.0}{20.9}{34.8} & \trip{2.6}{14.8}{36.5} & \trip{45.8}{85.6}{94.1} & \trip{35.0}{85.0}{90.0} \\
      GUIAct & \trip{47.4}{95.6}{98.4} & \trip{23.2}{60.5}{78.9} & \trip{25.0}{54.8}{59.6} & \trip{22.7}{57.5}{64.3} & \trip{91.9}{97.9}{99.3} & \trip{59.4}{85.9}{94.0} & \trip{16.2}{53.4}{59.4} & \trip{20.6}{57.5}{64.6} & \trip{34.6}{93.8}{98.4} & \trip{8.6}{41.9}{67.2} & \trip{41.8}{81.8}{89.7} & \trip{19.0}{44.0}{49.3} \\
      \bottomrule
    \end{tabular}
  \end{adjustbox}
  \label{tab:subtask_recall_ood}
\end{table}

\begin{table*}[h]
\scriptsize
\centering
\caption{Data annotation prompts.}

\begin{tabularx}{\textwidth}{lX}
    \toprule
    \multicolumn{1}{c}{\textbf{Function}} 
      & \multicolumn{1}{c}{\textbf{Content}} \\
    \midrule
    Description Generation
      & Generate a concise one-sentence description of the content and layout of the provided webpage screenshot. \\
    \midrule
    HTML Rendering
      & Your task is to convert the simplified HTML input provided by the user into a fully renderable, standard HTML format while preserving all original information intact. Enhance the HTML with appropriate styling to make it visually appealing and resemble a typical, functional website. Return only the HTML code without any additional text. HTML: \textit{html}. Ensure that the returned HTML code includes the ID [\textit{id}] (mentioned in \textit{context}) with the same element exactly as provided. \\
      \midrule
      Silver Generation
      & \textbf{Step 1}: You are a helpful AI assistant proficient in Named Entity Recognition (NER). Analyze the following sentence and provide the most comprehensive NER results for each noun in JSON format, using greedy matching. Labels should be specific contextual descriptions of the entity. Sentence: \textit{instruction}.\newline
        \textbf{Step 2}: You are an AI assistant skilled in generating alternatives. Given a sentence and a list of named entities, generate five alternative texts for each entity that align with its semantic label while being entirely different in meaning from the original text. Ensure the alternatives fit naturally and consistently within the sentence, maintaining the original representation (e.g., text remains text, emojis remain emojis). \newline Sentence: \textit{instruction}, Named Entities: \textit{ners}. \newline
        \textbf{Step 3}: You are an AI assistant specialized in rewriting user queries. Your task is to refine the following five queries to ensure they are consistent, natural, concise, logical, and human-like. Rewrite each query by varying the wording, structure, and style to ensure diversity in expression. Your response should align with real-world common sense and factual accuracy. \\
    \bottomrule
\end{tabularx}
\label{prompt:data_annotation}
\end{table*}

{\small
\begin{snippetbox}[colback=background_blue,colframe=frame_blue,rounded corners]{Action Space}{prompt:action_space}
\scriptsize{
\textbf{Mind2Web}
\begin{lstlisting}[mathescape=false, aboveskip=5pt, belowskip=5pt]
1. click: Simulates a mouse click on the target element (bounding box).
2. type: Types the specified value (str) into the target text input element (bounding box).
3. select: Selects the specified value (str) from a target dropdown element (bounding box).
\end{lstlisting}
\textbf{WebLINX}
\begin{lstlisting}[mathescape=false, aboveskip=5pt, belowskip=5pt]
1. click: Simulates a mouse click on the target element (bounding box).
2. hover: Simulates hovering over the target element (bounding box).
3. textInput: Types the value (str) into the target element (bounding box).
4. change: Changes the value of the target element (bounding box) to the specified value (str).
5. load: Loads the webpage at the specified url value (str).
6. submit: Submits the form identified by the target element (bounding box).
7. scroll: Scrolls the page to the specified coordinate values in the list of floats [x, y].
8. copy: Copies the specified text value (str) from the target element (bounding box).
9. paste: Pastes the specified text value (str) into the target element (bounding box).
\end{lstlisting}
\textbf{WebArena}
\begin{lstlisting}[mathescape=false, aboveskip=5pt, belowskip=5pt]
1. click: Simulates a mouse click on the target element (bounding box).
2. press: Simulates the pressing of a key combination value (str) on the target element (bounding box).
3. selectOption: Selects the specified option value (str) from the target dropdown element (bounding box).
4. check: Checks the target checkbox element (bounding box).
\end{lstlisting}
\textbf{GUIAct}
\begin{lstlisting}[mathescape=false, aboveskip=5pt, belowskip=5pt]
1. click: Clicks on the target element (bounding box).
2. hover: Hovers over the target element (bounding box).
3. input: Inputs the given text value (str) into the target element (bounding box).
4. scroll: Scrolls the screen by the values in the list of coordinate floats [down, right], where down represents vertical scroll and right represents horizontal scroll.
5. select_text: Selects text by dragging across the specified coordinate values in the list of floats [x1, y1, x2, y2], where (x1, y1) is the starting point and (x2, y2) is the ending point.
6. copy: Copies the specified text value (str) to the clipboard.
7. enter: Simulates pressing the Enter key.
8. select: Selects the text value (str) in the target element (bounding box).
9. answer: Provides an answer or response specified by text value (str) to the user.
\end{lstlisting}
\textbf{AutoWebGLM}
\begin{lstlisting}[mathescape=false, aboveskip=5pt, belowskip=5pt]
1. click: Clicks on the target element (bounding box).
2. hover: Hovers over the target element (bounding box).
3. select: Selects the option value (str) from a dropdown target element (bounding box).
4. type_string: Types the specified content (str) into the target element (bounding box) and presses Enter if press_enter (bool) is True. The action value is a list [content, press_enter].
5. scroll_page: Scrolls the page in the specified direction value ('up' or 'down').
6. go: Navigates browser history in the specified direction value ('forward' or 'backward').
7. jump_to: Opens the specified url (str) and optionally in a new tab if new_tab (bool) is True. The action value is a list [url, new_tab].
8. switch_tab: Switches to a browser tab specified by the value tab_index (int).
9. user_input: Displays the specified message (str) to obtain user input.
10. finish: Completes the task with an optional answer value (str or None).
\end{lstlisting}
}
\end{snippetbox}
}
{\small
\begin{snippetbox}[colback=background_gray,colframe=frame_gray,rounded corners]{Instruction Template}{prompt:instruction_template}
$(q, \tau_{1:i}) \to \tau_{i+1:n}$
\begin{lstlisting}[mathescape=true, basicstyle=\scriptsize, aboveskip=5pt, belowskip=5pt]
Determine the next web navigation trajectory using the task instruction "$\textit{description}$" and the prior trajectory below.
Retrieve the upcoming web navigation trajectory as specified by the task "$\textit{description}$" and the previous trajectory provided.
Search the next phase of the web navigation trajectory based on the user query "$\textit{description}$" and the earlier trajectory.
From the user input "$\textit{description}$" and the past navigation steps, locate the subsequent navigation sequence for GUI agents.
Apply the request "$\textit{description}$" to the previous web navigation steps to derive the next trajectory.
Represent the previous trajectory for web agents below to determine the next trajectory based on the task "$\textit{description}$".
According to the previous web agent trajectory below, identify the next sequence of steps to complete the user instruction "$\textit{description}$".
Based on the previous trajectory below, search the next interaction sequence of web agents for the user request "$\textit{description}$".
With the previous GUI navigation trajectory below as a guide, look for the next trajectory for the user intention "$\textit{description}$".
Given the goal "$\textit{description}$" and the earlier navigation sequence from GUI agents, match the subsequent trajectory.
\end{lstlisting}
$(q, \tau_{i+1:n}) \to \tau_{1:i}$
\begin{lstlisting}[mathescape=true, basicstyle=\scriptsize, aboveskip=5pt, belowskip=5pt]
Find the previous web browsing trajectory based on the user input "$\textit{description}$" and the current trajectory.
Identify the former web navigation history by analyzing the user request "$\textit{description}$" and the provided trajectory.
With the instruction "$\textit{description}$" and the following interaction sequence, extract the earlier trajectory for web agents.
Determine the past trajectory using the goal "$\textit{description}$" and the succeeding navigation sequence for GUI agents.
Retrieve the preceding GUI navigation trajectory with the task description "$\textit{description}$" and the succeeding trajectory below.
Using the user query "$\textit{description}$" and the succeeding navigation steps for web agents, locate the preceding interaction steps.
Analyze the query "$\textit{description}$" along with the provided web agent trajectory to derive the former navigation steps.
Based on the request "$\textit{description}$" and the later web interaction trajectory, look for the prior navigation sequence.
Consider the task "$\textit{description}$" together with the given trajectory to retrieve the prior web navigation trajectory.
Represent the current trajectory for web agents below to determine the previous trajectory according to the task "$\textit{description}$".
\end{lstlisting}
$(q, \tau_{1:i}) \to s_{i+1}$
\begin{lstlisting}[mathescape=true, basicstyle=\scriptsize, aboveskip=5pt, belowskip=5pt]
Identify the upcoming state from the earlier web navigation trajectory below and the instruction "$\textit{description}$".
Extract the next state from the provided previous navigation sequence for web agents and the directive "$\textit{description}$".
Locate the following state from the given former web navigation trajectory and the task input "$\textit{description}$".
Determine the subsequent observation from the task "$\textit{description}$" and the earlier web interaction trajectory.
Find the next observation by considering the command "$\textit{description}$" and the former GUI navigation trajectory.
Using the user instruction "$\textit{description}$" and the former web navigation trajectory, ascertain the subsequent state.
Based on the former web interaction history provided and the user request "$\textit{description}$", deduce the next state.
Use the user intention "$\textit{description}$" and the earlier navigation sequence for GUI agents to derive the following state.
Find the next state based on the goal "$\textit{description}$" and the earlier interaction history for web agents.
Represent the given GUI navigation history to locate the upcoming state according to the user intention "$\textit{description}$".
\end{lstlisting}
$(q, \tau_{i+1:n}) \to s_{i}$
\begin{lstlisting}[mathescape=true, basicstyle=\scriptsize, aboveskip=5pt, belowskip=5pt]
Find the antecedent state using the command "$\textit{description}$" and the current web navigation trajectory.
Identify the prior state by applying the directive "$\textit{description}$" along with the present navigation trajectory for GUI agents.
Locate the former observation using the instruction "$\textit{description}$" and the upcoming web interaction trajectory.
Ascertain the preceding observation based on the task input "$\textit{description}$" and the provided web agent browsing sequence.
Determine the previous state by employing the goal "$\textit{description}$" along with the upcoming navigation trajectory for web agents.
Find the antecedent state with the user instruction "$\textit{description}$" and the succeeding GUI navigation trajectory.
Retrieve the former observation given the request "$\textit{description}$" and the upcoming web navigation history.
Identify the prior state based on the task "$\textit{description}$" and the succeeding web interaction trajectory.
Represent the provided GUI navigation history to retrieve the previous state using the user query "$\textit{description}$".
Recognize the prior observation using the user task "$\textit{description}$" and the given web navigation trajectory.
\end{lstlisting}
$q \to \tau_{\equiv}$
\begin{lstlisting}[mathescape=true, basicstyle=\scriptsize, aboveskip=5pt, belowskip=5pt]
Determine the complete web navigation trajectory based on the following instruction "$\textit{description}$".
Locate the equivalent web navigation trajectory derived from the following user input "$\textit{description}$".
Find the GUI navigation history aligning with the following goal "$\textit{description}$".
Match the corresponding trajectory for web agents using the following user query "$\textit{description}$".
Pinpoint the equivalent navigation trajectory for GUI agents with the following task "$\textit{description}$".
Ascertain the corresponding web interaction trajectory using the request "$\textit{description}$".
Identify the unique navigation trajectory for web agents according to the provided instruction "$\textit{description}$".
Determine the complete GUI navigation trajectory from the user task "$\textit{description}$".
Ascertain the unique GUI interaction history by considering the task "$\textit{description}$".
Locate the exactly equivalent web navigation trajectory based on the given query "$\textit{description}$".
\end{lstlisting}
$q \to \tau_{\sim}$
\begin{lstlisting}[mathescape=true, basicstyle=\scriptsize, aboveskip=5pt, belowskip=5pt]
Determine the analogous web navigation trajectory based on the following directive "$\textit{description}$".
Identify the similar web navigation history using the task input "$\textit{description}$".
Locate the akin navigation sequence for GUI agents as dictated by the user input "$\textit{description}$".
Retrieve the similar web browsing trajectory as specified by the instruction "$\textit{description}$".
Identify the similar GUI interaction history based on the task description "$\textit{description}$".
Locate the analogous navigation trajectory for web agents using the instruction "$\textit{description}$".
Retrieve the analogous interaction history for GUI agents based on the provided command "$\textit{description}$".
Find a similar GUI navigation history following the task "$\textit{description}$".
Extract a similar web browsing trajectory based on the instruction "$\textit{description}$".
From the user query "$\textit{description}$", match the similar web agent navigation trajectory.
\end{lstlisting}
\end{snippetbox}

}

{\small
\begin{snippetbox}[colback=background_gray,colframe=frame_gray,rounded corners]{Instruction Template (Cont.)}{prompt:instruction_template_cont}
$(q, s_i) \to s_{i+1}$
\begin{lstlisting}[mathescape=true, basicstyle=\scriptsize, aboveskip=5pt, belowskip=5pt]
Retrieve the following web navigation state according to the instruction "$\textit{description}$" and the prior state.
Determine the subsequent web navigation observation given the task description "$\textit{description}$" and the preceding state.
Retrieve the upcoming observation for web navigation agents following the user input "$\textit{description}$" and the previous state.
Identify the next navigation state for GUI agents using the goal "$\textit{description}$" and the preceding state.
Using the provided instruction "$\textit{description}$" and the former state, what is the next GUI navigation state?
From the query "$\textit{description}$" and the prior observation, derive the next state in the web navigation trajectory.
Given the user input "$\textit{description}$" together with the current state, find the next web navigation state.
Taking the task input "$\textit{description}$" and the former state, what is the subsequent GUI navigation state?
Considering the directive "$\textit{description}$" and the preceding observation, determine the next web navigation state.
With the task "$\textit{description}$" and the previous state from web agents as inputs, determine the subsequent state.
\end{lstlisting}
$(q, s_{i+1}) \to s_i$
\begin{lstlisting}[mathescape=true, basicstyle=\scriptsize, aboveskip=5pt, belowskip=5pt]
Retrieve the prior web navigation state using the task "$\textit{description}$" along with the current state.
In light of the instruction "$\textit{description}$" and the current state provided, deduce the prior GUI navigation state.
Based on the provided user input "$\textit{description}$" and the current observation, find the prior navigation state for web agents.
With the directive "$\textit{description}$" and the current state, determine the prior web browsing state.
Considering both the current web agent observation provided and the user intention "$\textit{description}$", locate the prior navigation state.
Given the present state and the goal "$\textit{description}$", determine the previous GUI navigation state.
Combine the task description "$\textit{description}$" with the current state to identify the preceding navigation state for GUI agents.
Taking the description "$\textit{description}$" and the current state into account, search the previous web agent state.
Utilize the user request "$\textit{description}$" alongside the present state to extract the prior GUI navigation state.
Use the directive "$\textit{description}$" with the current state to look for the state directly preceding in the web agent navigation trajectory.
\end{lstlisting}
$(q, s_i) \to \tau_{i+1:n}$
\begin{lstlisting}[mathescape=true, basicstyle=\scriptsize, aboveskip=5pt, belowskip=5pt]
Find the subsequent web navigation trajectory based on the instruction "$\textit{description}$" and the previous state.
Based on the task "$\textit{description}$" and the previous observation, identify the subsequent GUI navigation trajectory.
Locate the next GUI navigation trajectory by applying the instruction "$\textit{description}$" to the previous state.
With the user input "$\textit{description}$" and the previous state in hand, identify the next navigation trajectory for web agents.
What is the following navigation trajectory for GUI agents when applying the user intention "$\textit{description}$" to the previous state?
Given the user goal "$\textit{description}$" and the previous state, search the next web navigation trajectory.
When given the user instruction "$\textit{description}$" and the former state, what is the next trajectory for web navigation?
Identify the next web navigation trajectory by merging the task "$\textit{description}$" with the previous state.
From the directive "$\textit{description}$" and the prior state, look for the subsequent GUI navigation trajectory.
Determine the subsequent browsing trajectory for web agents with the task "$\textit{description}$" and the previous state as references.
\end{lstlisting}
$(q, s_{i+1}) \to \tau_{1:i}$
\begin{lstlisting}[mathescape=true, basicstyle=\scriptsize, aboveskip=5pt, belowskip=5pt]
Find the previous web navigation history based on the instruction "$\textit{description}$" and the current state.
Retrieve the preceding web navigation trajectory using the intention "$\textit{description}$" along with the present state.
From the instruction "$\textit{description}$" and the present state, find the prior GUI navigation history.
What does the previous navigation history for web agents look like when derived from the user input "$\textit{description}$" and the current state?
Locate the prior GUI navigation history by combining the description "$\textit{description}$" with the current observation.
Identify the web navigation trajectory preceding the current state according to the task "$\textit{description}$".
Derive the previous navigation trajectory for GUI agents by combining the instruction "$\textit{description}$" with the current state.
Search the browsing history for web agents that came before the provided current observation with regard to the user query "$\textit{description}$".
Based on the current state and the user intention "$\textit{description}$", extract the trajectory that came before in the web navigation.
Recognize the GUI navigation history that predates the current state by considering the user need "$\textit{description}$".
\end{lstlisting}
$q \to s_i$
\begin{lstlisting}[mathescape=true, basicstyle=\scriptsize, aboveskip=5pt, belowskip=5pt]
Find the specific web navigation state corresponding to the description "$\textit{description}$".
Identify the equivalent web navigation state as defined by the description "$\textit{description}$".
Extract the GUI navigation observation that corresponds with "$\textit{description}$".
Locate the web navigation state as dictated by the description "$\textit{description}$".
Identify the navigation state for GUI agents that is equivalent to the details provided in "$\textit{description}$".
Search the observation for web navigation that best fits the details described in "$\textit{description}$".
Determine the precise GUI navigation observation that reflects the input "$\textit{description}$".
Retrieve the navigation observation for web agents that best aligns with the input "$\textit{description}$".
What is the corresponding web browsing state described by the input "$\textit{description}$"?
From the description "$\textit{description}$", identify the specific navigation observation for GUI navigation.
\end{lstlisting}
$q \to s_n$
\begin{lstlisting}[mathescape=true, basicstyle=\scriptsize, aboveskip=5pt, belowskip=5pt]
Retrieve the last web navigation observation based on the task "$\textit{description}$".
From the task "$\textit{description}$", locate the final state in the web navigation sequence.
What is the ultimate web navigation state for the task instruction "$\textit{description}$"?
Find the end state in the GUI navigation as defined by the task "$\textit{description}$".
Determine the last state in the web browsing process for the task "$\textit{description}$".
Locate the final observation of the web navigation for the task "$\textit{description}$".
Identify the concluding status in the GUI agent trajectory for the task "$\textit{description}$".
Based on the task "$\textit{description}$", extract the final navigation observation for web agents.
What is the final GUI navigation status according to the task "$\textit{description}$"?
Search the terminal observation in the web navigation for the task instruction "$\textit{description}$".
\end{lstlisting}
\end{snippetbox}

}

    \end{document}